\documentclass[10pt, a4paper]{article}

\usepackage[final]{lrec2026} % this is the new style
\usepackage{inconsolata}
\usepackage{booktabs}
\usepackage{soul}
\usepackage{color}

\usepackage{multirow}
\usepackage{multicol}
\usepackage{enumitem,kantlipsum}
\usepackage[normalem]{ulem}
\usepackage{color,colortbl}
\usepackage{todonotes}
\usepackage{soul}
\usepackage{threeparttable, makecell}
\usepackage{makecell}
\usepackage{xcolor}

\usepackage{tcolorbox}

\usepackage{comment}
\usepackage{array}
\usepackage{lscape}
\usepackage{multirow}
\usepackage{graphicx}
\usepackage[normalem]{ulem}
\useunder{\uline}{\ul}{}

\usepackage{array}
\usepackage{multirow}
\usepackage{algorithm}
\usepackage{algpseudocode}
\usepackage{amsmath}
\usepackage{subcaption}
\usepackage{fancybox}
\usepackage{tikz}
\usepackage{listings}
\usepackage{color}
\usepackage{pifont} 

\usepackage{enumitem}
\setlist{nosep,topsep=-\parskip}

\usepackage{soul}
\usepackage{xcolor}
\definecolor{dkgreen}{rgb}{0,0.6,0}

\usepackage{listings}
\usepackage{graphicx}
\definecolor{codegray}{gray}{0.9}
\definecolor{ftblue}{RGB}{230,240,255}
\definecolor{gptgray}{RGB}{245,245,245}

\definecolor{arabicavg}{HTML}{E8F6F3} 

\lstdefinestyle{pythonstyle}{
    backgroundcolor=\color{codegray},   
    language=Python,
    basicstyle=\ttfamily\footnotesize,
    keywordstyle=\color{blue},
    stringstyle=\color{red},    
    breaklines=true,
    frame=single,
    keepspaces=true,
    showstringspaces=false,
}

\usepackage{url}
\usepackage{xcolor}
\usepackage{soul}
\sethlcolor{lightblue}

\usepackage{graphicx}
\definecolor{lightblue}{rgb}{.50,.95,1}
\definecolor{tri}{rgb}{.25,.88,.82}
\definecolor{lilac}{rgb}{0.85,0.64,0.85}

\usepackage{amssymb}
\usepackage{hyperref}
\usepackage{verbatim}
\usepackage{setspace}

\usepackage{colortbl}
\usepackage{booktabs}
\usepackage{graphicx}
\usepackage{mathrsfs}
\usepackage{xcolor}
\definecolor{mycolor}{RGB}{0, 110, 180}
\usepackage{upquote}

\title{Beyond MCQ: An Open-Ended Arabic Cultural QA Benchmark with Dialect Variants}

\name{Hunzalah Hassan Bhatti, Firoj Alam}

\address{Qatar Computing Research Institute, Qatar \\
         fialam@hbku.edu.qa, hunzalahhassan@gmail.com
}

\abstract{
Large Language Models (LLMs) are increasingly used to answer everyday questions, yet their performance on culturally grounded and dialectal content remains limited across languages and their varieties. We propose a comprehensive method that \textit{(i)} translates Modern Standard Arabic (MSA) multiple-choice questions (MCQs) into English and several Arabic dialects, \textit{(ii)} converts them into open-ended questions (OEQs), \textit{(iii)} benchmarks a range of zero-shot and fine-tuned LLMs under both MCQ and OEQ settings, and \textit{(iv)} generates chain-of-thought (CoT) rationales to fine-tune models for step-by-step reasoning. Using this method, we extend an existing dataset in which QAs are parallelly aligned across language varieties, making it, to our knowledge, the \textit{first} of its kind. A large portion of the resulting test set is further validated through targeted human annotation and native-speaker post-editing. We conduct extensive experiments with both open and closed models. Our findings show that \textit{(i)} models underperform on Arabic dialects, showing persistent gaps in culturally grounded and dialect-specific knowledge; \textit{(ii)} Arabic-centric models perform well on MCQs but struggle with OEQs; and \textit{(iii)} CoT improves judged correctness while yielding mixed n-gram-based metrics.
\\ \newline 
}

\begin{document}

\maketitleabstract

% \section{Introduction}
% Please use the \texttt{[review]} setting for submissions:

\section{Introduction}
\label{sec:introduction}

Cultural information underpins human identity, behavior, and social interaction, encompassing shared beliefs, values, customs, languages, traditions, and collective practices. In today's tightly coupled information-communication ecosystem, hundreds of millions of users interact with LLMs for everyday queries, often asking about local norms, holidays, cuisine, or etiquette, where culturally grounded interpretations are essential \cite{pawar2024survey,hasan-etal-2025-nativqa}. Yet despite rapid progress in multilingual understanding and reasoning, LLM performance remains uneven across languages, dialects, and culturally specific domains \citep{weichain,muennighoff-etal-2023-crosslingual}. The issue is especially salient for Arabic, where MSA coexists with numerous regional dialects that differ in phonology, morphology, lexicon, and usage \citep{alwajih-etal-2025-palm,sadallah-etal-2025-commonsense}. Beyond modeling challenges, widely used MCQ evaluations can mask deficiencies in reasoning by enabling superficial answer-selection strategies such as label bias or option-guessing, complicating fair cross-lingual and cross-format comparison \citep{raman2025reasoning,li2024can}. 

\begin{figure}
\centering
\includegraphics[width=1.0\linewidth]{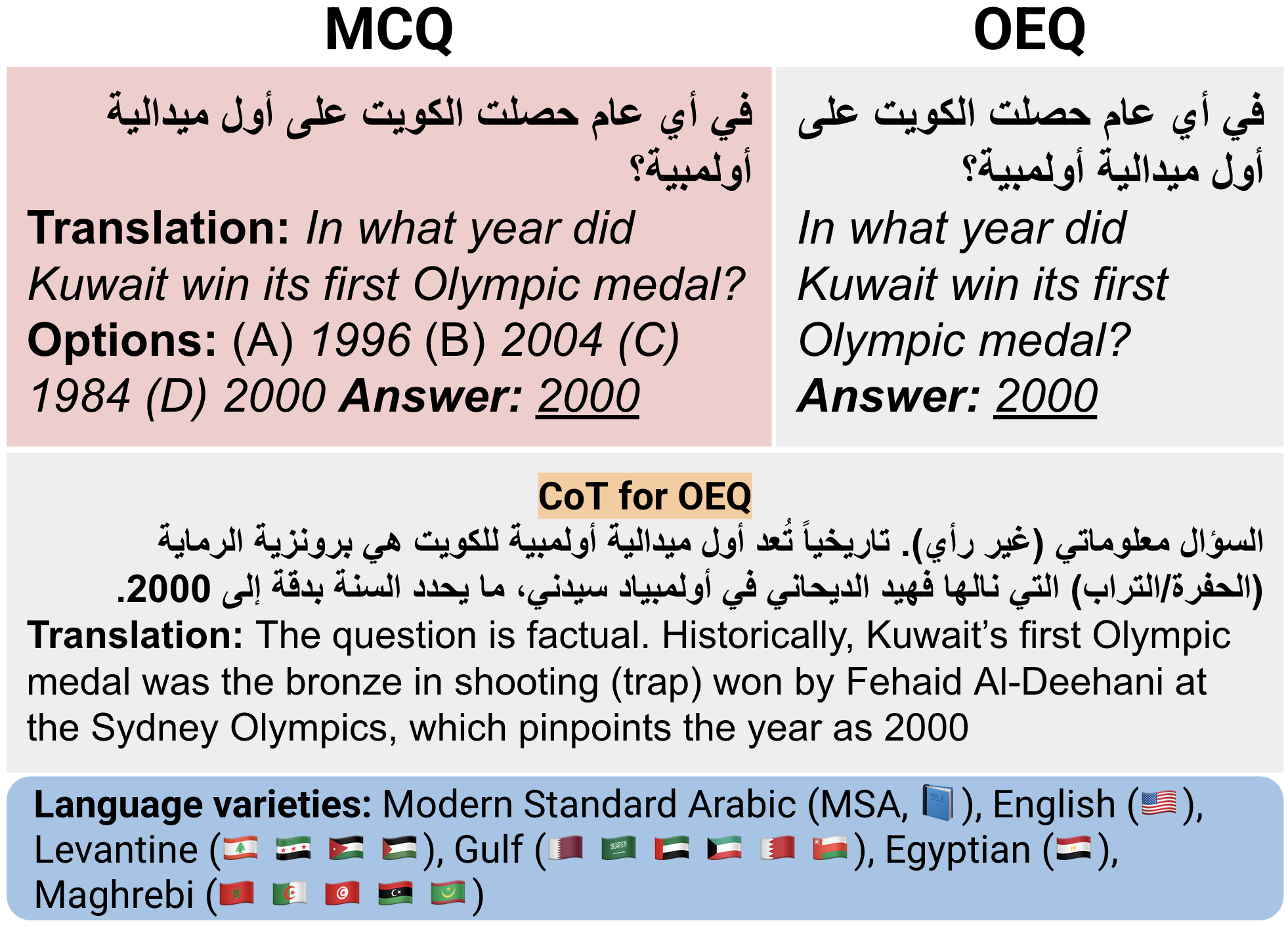}
% \vspace{-0.4cm}
\caption{Example QA instances shown in two formats: multiple-choice question (MCQ) and open-ended question (OEQ). Flags in parentheses indicate representative countries where each dialect is widely spoken.
% shown in two formats (MCQ and OEQ). MCQ: Multiple-Choice Question; OEQ: Open-Ended Question. Flags in parentheses indicate representative countries where each dialect is widely spoken.
}
% \vspace{-0.4cm}
\label{fig:palmx_example}
% \vspace{-0.3cm}
\end{figure}

A central open problem is how to \emph{measure} and \emph{improve} an LLM's ability to understand and generate responses to such culturally embedded queries, especially in multilingual settings with substantial dialectal variation. Another noteworthy aspect is that MCQs have long been the dominant format for evaluating QA performance in LLMs due to their simplicity, automatic scoring, and structured answer space \cite{myrzakhan2024open}. However, models can sometimes exploit the test format rather than genuinely understanding the question, leading to a form of selection bias, for instance, consistently favoring certain options (e.g., always choosing ``A'') regardless of content.

To address these challenges, parallel efforts have emerged to develop culturally aligned language models \cite{wang2023aligning} and to enable their efficient deployment in low-compute environments \cite{hu2022lora}. At the same time, new culturally relevant datasets, targeted benchmarks, and evaluation protocols are beginning to operationalize the measurement of everyday cultural knowledge \cite{myung2024blend,NEURIPS2024_77f089cd,mousi-etal-2025-aradice,alam2025nativqaframeworkenablingllms,alam2025everydaymmqa}. Collectively, these trends demonstrate the need for new resources, evaluations, and models that are grounded in underrepresented dialectal varieties and culturally contextualized content.

% \textcolor{red}{maybe need to add a little bit more about the manual checking? or not?}
To shade a light on the challenges, we introduce a comprehensive method for developing a new resource for under-representative language verities. Starting from an existing MSA MCQ dataset \cite{alwajih2025palmx}, we perform the following steps: \textit{(i)} translate the questions into several Arabic dialects and English, which were then manually post-edited \textit{(ii)} convert the MCQs into OEQ that require free-form answers, \textit{(iii)} evaluate a range of zero-shot and fine-tuned LLMs on the resulting benchmark, and \textit{(iv)} create and fine-tune models on chain-of-thought (CoT) annotations to encourage explicit reasoning for OEQ. An example of MCQ, OEQ with CoT is shown in Figure \ref{fig:palmx_example}.

Our approach allows us to isolate and study the impact of question format, language variety, and reasoning supervision on model performance. 
% \textcolor{red}{
We find that OEQ settings present greater challenges than MCQ, especially in dialectal Arabic.
% , and that fine-tuning on CoT data improves both accuracy and interpretability. 
% Our benchmark surfaces specific gaps in everyday knowledge and cultural understanding, which are often obscured in standard QA datasets.
% }
Our contributions are as follows:
\begin{itemize}
% [noitemsep,topsep=0pt,leftmargin=*,labelsep=.5em] 
    \item We construct a multilingual and multidialectal QA dataset, \textit{\textbf{ArabicCulturalQA}},  by translating MSA MCQs into English and Arabic dialects. The dataset is publicly available for research use.\footnote{\href{https://huggingface.co/datasets/QCRI/ArabicCulturalQA}{QCRI/ArabicCulturalQA}}
    \item We convert the dataset into OEQs in all language variants, enabling a more rigorous evaluation of model knowledge.
    \item A substantial portion of the test set is human annotated by native speakers: dialectal MCQs are post-edited, and the conversion from MSA MCQs to MSA OEQs is manually reviewed to ensure linguistic and semantic fidelity.
    \item We benchmark a range of zero-shot and fine-tuned LLMs under both MCQ and OEQ settings.
    \item We generate chain-of-thought (CoT) annotations for OEQ and fine-tune models.
    % to perform step-by-step reasoning.
\end{itemize}

This work represents the \textit{first} effort to unify dialectal Arabic QA, open-ended reasoning, and CoT fine-tuning in a single benchmark, offering new insights into LLM performance on culturally rich, linguistically diverse data.

\section{Related Work}
\label{sec:relatedwork}

\noindent
\subsection{General Capabilities of LLMs.}
% LLMs have demonstrated impressive general capabilities across a variety of NLP tasks, including text generation, translation, summarization, and reasoning ~\cite{abdelali-etal-2024-larabench}. 
LLMs have shown strong generalization across a broad range of NLP tasks, including text generation, translation, summarization, and reasoning~\cite{abdelali-etal-2024-larabench}.
At sufficient scale, LLMs exhibit \textit{emergent abilities}, such as multi-step inference and commonsense reasoning \cite{bubeck2023sparks,weichain}. Prompting techniques like few-shot and chain-of-thought (CoT) significantly enhance performance on reasoning-heavy tasks \cite{kojima2022large,weichain}. However, most evaluations focus on English or high-resource languages. Performance often degrades on morphologically rich or low-resource languages such as Arabic, particularly in dialectal contexts \cite{mousi-etal-2025-aradice,muennighoff-etal-2023-crosslingual}. 

\noindent
\subsection{Cultural and Everyday Knowledge.}
Recent research has highlighted the limitations of LLMs in capturing culturally grounded, everyday knowledge. Myung et al.~\shortcite{myung2024blend} introduced BLEnD, a multilingual benchmark comprising 52.6K QA pairs across 13 languages and 16 regions, designed to evaluate models’ understanding of daily-life knowledge. Similarly, Hasan et al.~\cite{hasan-etal-2025-nativqa} developed MultiNativQA, featuring 64K QA pairs covering nine locations in seven languages. Across these studies, results consistently show that LLMs underperform on questions reflecting underrepresented cultures, often reflecting Western-centric norms. In the Arabic context, \citet{sadallah-etal-2025-commonsense} proposed \textsc{ArabCulture}, a benchmark of 3.5K MSA-based MCQs curated  by native speakers from 13 Arab countries to assess culturally specific commonsense reasoning. Likewise, \citet{alwajih-etal-2025-palm} introduced \textsc{Palm}, a dialect-rich dataset encompassing all 22 Arab countries. 

\noindent
% \subsection{Challenges in Converting MCQ to OEQ.}
\subsection{MCQ to OEQ.}
Many evaluation benchmarks use MCQs because they allow straightforward automatic scoring, in which the model selects an option (A/B/C/D) that can be directly compared with the correct answer. However, recent studies show that this format may introduce artificial performance gains and {mask a model’s actual reasoning ability~\cite{molfese2025right,chandak2025answer,myrzakhan2024open}. For instance, LLMs often display a \textit{selection bias}, favoring certain options (e.g., consistently choosing ``A'') due to training artifacts. To mitigate these issues, several works propose converting MCQs into OEQs that require the model to generate answers without predefined choices~\cite{myrzakhan2024open}. This forces reliance on internal knowledge and reasoning rather than elimination or guessing. Yet, this conversion introduces new challenges: some MCQs become ambiguous once options are removed, and others may yield multiple valid answers unless carefully rephrased. Moreover, evaluating free-form responses is inherently harder, as correctness depends on comparing generated text with gold answers that may differ in wording. Prior work addresses this by using LLM-based evaluation pipelines (e.g., GPT-4) to judge open-ended answers against human references with high reliability~\cite{myrzakhan2024open}. Overall, shifting from MCQ to open-ended formats holds promise for revealing deeper model understanding, but it demands careful question selection and robust evaluation protocols.

\noindent
\subsection{Chain-of-Thought (CoT) Reasoning.}
CoT prompting has emerged as a powerful technique for enhancing reasoning capabilities in LLMs. Instead of producing an answer directly, the model is encouraged to generate an explicit, step-by-step reasoning path before reaching a final conclusion~\cite{weichain}. By articulating these intermediate steps, models can decompose complex problems into manageable components, leading to substantial gains in accuracy.  Remarkably, even without task-specific training, simply prefixing the prompt with \textit{``Let's think step by step''} can induce this behavior in sufficiently large models, a method known as \textit{zero-shot CoT}~\cite{qin-etal-2023-cross}. This simple prompting strategy has demonstrated significant improvements across a wide range of reasoning tasks, including mathematical problem solving and commonsense reasoning. Furthermore, \citet{qin-etal-2023-cross} introduced a \textit{self-consistency} mechanism, in which the model generates multiple reasoning chains and selects the most frequent answer, further enhancing performance. While most existing studies emphasize inference-time CoT, recent research has explored \textit{CoT fine-tuning} to transfer reasoning skills to smaller or multilingual models~\cite{puerto-etal-2025-fine}. However, to the best of our knowledge, no prior work has applied CoT fine-tuning to Arabic open-ended QA datasets, particularly those covering dialectal varieties, which constitutes a key contribution of our study.

\section{Datasets}
\label{sec:dataset}

% Our data is based on the \textbf{PalmX 2025 - General Culture Evaluation (PalmX-GC)} dataset, which assesses a model's understanding of Arab culture, including customs, history, geography, arts, cuisine, notable figures, and everyday life across the 22 Arab countries. All questions and answers are written in MSA and manually verified, providing a high quality benchmark for culturally grounded QA~\cite{alwajih2025palmx}. The dataset comprises 2,000 training, 500 development, and 2000 test examples, all in MCQ format. We use PalmX-GC as the foundation for creating dialectal MCQ and OEQ variants. Figure~\ref{fig:dataset_pipeline} illustrates the dataset construction process, in which, we used LLMs (specifically GPT-4.1) for both translation and data conversion. Our choice of this model was primarily based on its reliability and our available paid access. 

Our data, \textbf{\textit{ArabicCulturalQA}}, is based on the \textbf{PalmX 2025 - General Culture Evaluation (PalmX-GC)} dataset, which assesses a model's understanding of Arab culture, including customs, history, geography, arts, cuisine, notable figures, and everyday life across the 22 Arab countries. All questions and answers are written in \textit{MSA} and \textit{Manually Verified}, providing a high-quality benchmark for culturally grounded QA~\cite{alwajih2025palmx}. The dataset comprises 2,000 training, 500 development, and 2,000 test examples, all in MCQ format. We use PalmX-GC as the basis for creating dialectal MCQ and OEQ variants. Figure~\ref{fig:dataset_pipeline} illustrates the dataset construction process, in which we used LLMs (specifically GPT-4.1) for translation and data conversion. We chose this model based on its reliability and our paid access.

\begin{figure*}
\centering
\includegraphics[width=0.65\linewidth]{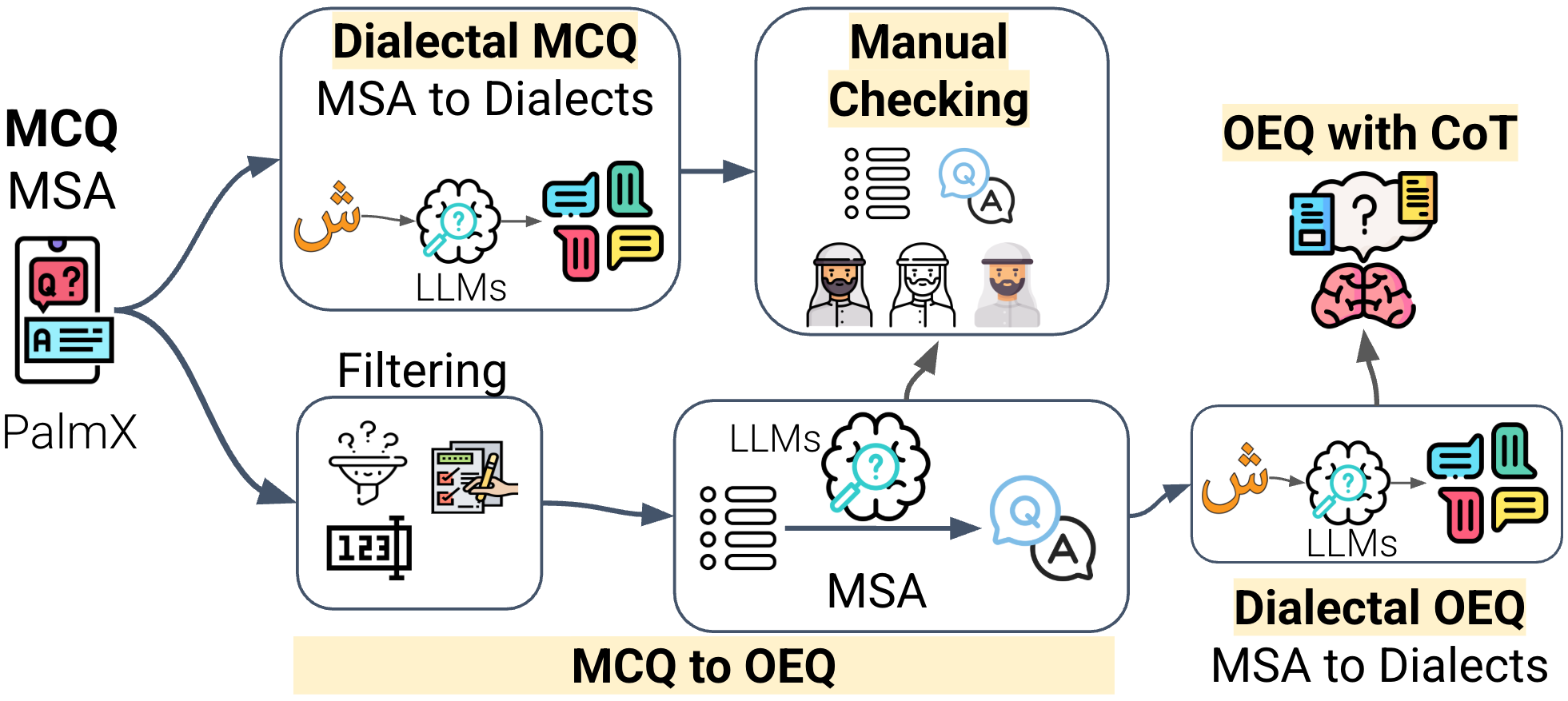}
% \vspace{-0.3cm}
\caption{Pipeline for the dataset construction process.}
\label{fig:dataset_pipeline}
% \vspace{-0.5cm}
\end{figure*}

\subsection{Dialectal MCQ}
% To broaden cultural and linguistic coverage beyond MSA, we translate PalmX into four Arabic dialect  Egyptian, Levantine, Gulf, and Maghrebi, and into English GPT-4.1 with a quality check. These dialects were selected because \textit{(i)} they collectively cover the largest speaker populations and the widest geographic spread across the Arab world; \textit{(ii)} they capture major points on the Arabic dialect continuum with substantial lexical, morphological, and pragmatic divergence from MSA; and \textit{(iii)} they are the primary medium of everyday communication and online discourse, where culturally grounded queries naturally occur. Including English serves two purposes: it provides a shared reference baseline for cross-lingual comparison (disentangling language modeling from culture-specific knowledge) and reflects real usage, where users often pose culturally focused questions in English about Arabic contexts. This design enables us to probe \textit{(a)} format sensitivity (MCQ$\rightarrow$OEQ), \textit{(b)} dialect sensitivity (MSA vs.\ regional varieties), and \textit{(c)} cross-lingual transfer (Arabic$\leftrightarrow$English) within a single, controlled benchmark.

To broaden cultural and linguistic coverage beyond MSA, we translate PalmX into four Arabic dialects such as Egyptian, Levantine, Gulf, and Maghrebi and into English using GPT-4.1, followed by quality checking. We selected these dialects because \textit{(i)} they cover the largest speaker populations and broadest geographic span in the Arab world, \textit{(ii)} capture major points on the Arabic dialect continuum, and \textit{(iii)} represent the main language of everyday communication and online discourse. Including English serves two purposes: it provides a shared reference baseline for cross-lingual comparison, helping disentangle language modeling from culture-specific knowledge, and reflects real usage, where users often ask culturally grounded questions in English about Arabic contexts. This design allows us to probe \textit{(a)} format sensitivity (MCQ$\rightarrow$OEQ), \textit{(b)} dialect sensitivity (MSA vs.\ regional varieties), and \textit{(c)} cross-lingual transfer (Arabic$\leftrightarrow$English) within a single controlled benchmark.

We employed controlled prompting to translate each MSA MCQ into four dialects and \textit{English}. The prompts explicitly enforced semantic equivalence while allowing lexical and stylistic adaptation to dialectal norms. This approach ensured that the dialectal phrasing preserved the original question’s intent without causing any semantic drift from its MSA counterpart.

\subsection{MCQ to OEQ}
We converted the MSA MCQs into OEQs using \texttt{GPT-4.1}. Each MCQ was transformed into a natural QA pair by rephrasing the original question and its correct option into a single, self-contained QA instance. The remaining distractors were used only to guide contextual understanding but were excluded from the final prompt. We filtered out QA items where conversion was structurally infeasible, such as questions dependent on visible alternatives, to avoid ill-posed or underspecified open-ended forms. This process ensured that the resulting OEQs were faithful derivations of verified MCQs rather than arbitrary generations.

% We converted the MSA MCQs into OEQs using \texttt{GPT-4.1}. Each MCQ is converted into a natural QA pair by rephrasing the original question and its correct choice into a single self-contained QA instance. The remaining options are used solely to guide contextual understanding but are not part of the final prompt. We filter out cases where conversion is structurally impossible, such as numeric selections or questions that depend on visible alternatives, to avoid ill-posed or underspecified open-ended forms. This ensures that the resulting OEQs are faithful derivations of verified MCQs rather than arbitrary generations.

\subsection{Dialectal OEQ}
% We then translated the OEQs into dialectal variants using similar controlled prompting principles, promoting authentic dialectal translation while preserving both semantic and pragmatic fidelity to the original MSA version. The resulting dataset constitutes a parallel corpus spanning five dialects and English, each aligned with consistent cultural grounding and verified equivalence. This parallel structure enables systematic evaluation of dialectal reasoning and transfer capabilities in generative settings.
We then translated the OEQs into dialectal variants using similar controlled prompting, encouraging natural dialectal expression while preserving the semantic and pragmatic meaning of the original MSA version. The resulting dataset forms a parallel corpus across five Arabic varieties and English, aligned around the same cultural content and verified for equivalence. This structure enables systematic evaluation of dialectal reasoning and cross-variant transfer in generative settings.

\subsection{OEQ with CoT}

Inspired by prior work \cite{yu2025cotselfinstruct,zelikman2022star}, we transform each OEQ instance \(x = (q, a^\star)\) $\in$ dataset $D$, where \(q\) denotes the question and \(a^\star\) the gold or reference answer, into one or more CoT training samples using a four-stage pipeline. The pipeline generates multiple reasoning chains without revealing the gold answer, optionally produces gold-conditioned rationalizations, and verifies accepted chains. While generating CoTs, we also prompt the LLM to classify each \(q\) as either \textit{factual} or \textit{subjective}. Identifying the question type enables type-specific model development and evaluation. For instance, factual questions may require reference evidence or source attribution for their answers.

\paragraph{Preliminaries.}
Let $N=\texttt{samples}$ be the number of chain attempts, $T=\texttt{rationalize\_target}$ the minimum number of gold-aligned chains to retain, and $\rho=\texttt{accept\_ratio}\in(0,1]$ the acceptance threshold. For each attempt $i\in\{1,\dots,N\}$, we obtain $(\hat{c}_i,\hat{a}_i,\hat{\ell}_i)$, where $\hat{c}_i\in\mathcal{C}$ is a generated chain-of-thought, $\hat{a}_i\in\mathcal{Y}$ is the generated answer, and $\hat{\ell}_i\in\mathcal{L}=\{\textsf{factual},\textsf{subjective}\}$ is a label. Let also denote the collection of attempts as
\[
\mathcal{S}=\{(\hat{c}_i,\hat{a}_i,\hat{\ell}_i)\}_{i=1}^N,\qquad
\mathcal{K}\subseteq\mathcal{S}~(accepted~subset).
\]
Acceptance is determined via a matching function $\textsf{match}:\mathcal{Y}\times\mathcal{Y}\to\{0,1\}$ (see \textsc{Match}) against the gold answer $a^\star$, with indicator
\[
m_i=\mathbb{I}\{\textsf{match}(\hat{a}_i,a^\star)=1\}.
\]
We enforce $\lvert\mathcal{K}\rvert\ge T$ and the empirical acceptance ratio $\lvert\mathcal{K}\rvert/N\ge\rho$.

% \paragraph{Preliminaries.}
% Let $N=\texttt{samples}$ be the number of chain attempts, $T=\texttt{rationalize\_target}$ the minimum number of gold-aligned chains to retain, and $\rho=\texttt{accept\_ratio}\in(0,1]$ the acceptance threshold. For each attempt $i\in\{1,\dots,N\}$, we obtain $(\hat{c}_i,\hat{a}_i,\hat{\ell}_i)$, where $\hat{c}_i\in\mathcal{C}$ is a generated chain-of-thought, $\hat{a}_i\in\mathcal{Y}$ is the generated answer, and $\hat{\ell}_i\in\mathcal{L}=\{\textsf{factual},\textsf{subjective}\}$ is a label. Let also denote the collection of attempts as $\mathcal{S}=\{(\hat{c}_i,\hat{a}_i,\hat{\ell}_i)\}_{i=1}^N$ and the accepted subset as $\mathcal{K}\subseteq\mathcal{S}$. Acceptance is determined via a matching function $\textsf{match}:\mathcal{Y}\times\mathcal{Y}\to\{0,1\}$ (see \textsc{Match}) against the gold answer $a^\star$, with indicator $m_i=\mathbb{I}\{\textsf{match}(\hat{a}_i,a^\star)=1\}$. We enforce $\lvert\mathcal{K}\rvert\ge T$ and the empirical acceptance ratio $\lvert\mathcal{K}\rvert/N\ge\rho$.

% \textbf{1.\ CoT generation.}
% Let $\mathcal{G}:\mathcal{Q}\to\mathcal{C}\times\mathcal{Y}\times\mathcal{L}$ denote the rationale-answer-label generator. For each $q\in\mathcal{Q}$, we sample
% $(\hat{c}_i,\hat{a}_i,\hat{\ell}_i)=\mathcal{G}(q)$.
% For each attempt, compute the match flag $m_i=\mathbb{I}\{\textsf{match}(\hat{a}_i,a^\star)=1\}$ and collect the kept subset
% $\mathcal{K}=\{(\hat{c}_i,\hat{a}_i,\hat{\ell}_i)\in\mathcal{S}: m_i=1\}$.
\vspace{0.3cm}
\noindent
\textbf{1.\ CoT generation.}
Let $\mathcal{G}:\mathcal{Q}\to\mathcal{C}\times\mathcal{Y}\times\mathcal{L}$ denote the rationale-answer-label generator. For each $q\in\mathcal{Q}$, we sample $(\hat{c}_i,\hat{a}_i,\hat{\ell}_i)=\mathcal{G}(q)$. For each attempt, compute the match flag
\[
m_i=\mathbb{I}\{\textsf{match}(\hat{a}_i,a^\star)=1\},
\]
and collect the kept subset
\[
\mathcal{K}=\{(\hat{c}_i,\hat{a}_i,\hat{\ell}_i)\in\mathcal{S}: m_i=1\}.
\]

% \textbf{2.\ CoT rationalize with gold.}
% Let $\mathcal{R}:\mathcal{Q}\times\mathcal{Y}\to\mathcal{C}\times\mathcal{Y}\times\mathcal{L}$ denote the gold-conditioned rationalizer.
% If $|\mathcal{K}|<T$ (as obtained in Step~1), we draw additional chains via
% $(\tilde{c}_j,\tilde{a}_j,\tilde{\ell}_j)=\mathcal{R}(q,a^\star)$
% and retain those that match the gold answer:
% $\tilde{m}_j=\mathbb{I}\{\textsf{match}(\tilde{a}_j,a^\star)=1\}$,
% $\widetilde{\mathcal{K}}=\{(\tilde{c}_j,\tilde{a}_j,\tilde{\ell}_j):\tilde{m}_j=1\}$.
% We then update $\mathcal{K}\leftarrow \mathcal{K}\cup\widetilde{\mathcal{K}}$
% until $|\mathcal{K}|\ge T$ (and, if applicable, $|\mathcal{K}|/N\ge\rho$).
% This stage ensures a sufficient pool of gold-aligned CoT for downstream use.

\vspace{0.3cm}
\noindent
\textbf{2.\ CoT rationalize with gold.}
Let $\mathcal{R}:\mathcal{Q}\times\mathcal{Y}\to\mathcal{C}\times\mathcal{Y}\times\mathcal{L}$ denote the gold-conditioned rationalizer. If $|\mathcal{K}|<T$ (as obtained in Step~1), we draw additional chains via $(\tilde{c}_j,\tilde{a}_j,\tilde{\ell}_j)=\mathcal{R}(q,a^\star)$ and retain those that match the gold answer, with
\[
\tilde{m}_j=\mathbb{I}\{\textsf{match}(\tilde{a}_j,a^\star)=1\},
\]
\[
\widetilde{\mathcal{K}}=\{(\tilde{c}_j,\tilde{a}_j,\tilde{\ell}_j):\tilde{m}_j=1\}.
\]
We then update $\mathcal{K}\leftarrow \mathcal{K}\cup\widetilde{\mathcal{K}}$ until $|\mathcal{K}|\ge T$ (and, if applicable, $|\mathcal{K}|/N\ge\rho$). This stage ensures a sufficient pool of gold-aligned CoT for downstream task.

% \textbf{3.\ Verification.}
% Let $\mathcal{V}:\mathcal{Q}\times\mathcal{Y}\times\mathcal{C}\times\mathcal{Y}\to[0,1]\times\{\textsf{pass},\textsf{fail}\}\times\mathsf{Report}$ 
% denote the verifier, where the two $\mathcal{Y}$ components correspond to the gold answer $a^\star$ and the candidate answer $a_k$, respectively.
% For each retained item $(c_k,a_k,\ell_k)\in\mathcal{K}$, compute
% $(\sigma_k,\nu_k,r_k)=\mathcal{V}(q,a^\star,c_k,a_k)$,
% where $\sigma_k\in[0,1]$ is a confidence score, $\nu_k\in\{\textsf{pass},\textsf{fail}\}$ is the verdict under a default threshold $\tau=0.8$ (i.e., $\nu_k=\textsf{pass}$ iff $\sigma_k\ge\tau$), and $r_k$ is a brief issue report.
% We then form the verified subset
% $\mathcal{K}_{\mathrm{ver}}=\{(c_k,a_k,\ell_k)\in\mathcal{K}:\nu_k=\textsf{pass}\}$,
% noting that $\ell_k$ is carried forward but not used by $\mathcal{V}$.

\vspace{0.3cm}
\noindent
\textbf{3.\ Verification.}
Let $\mathcal{V}:\mathcal{Q}\times\mathcal{Y}\times\mathcal{C}\times\mathcal{Y}\to[0,1]\times\{\textsf{pass},\textsf{fail}\}\times\mathsf{Report}$ denote the verifier, where the two $\mathcal{Y}$ components correspond to the gold answer $a^\star$ and the candidate answer $a_k$, respectively. For each retained item $(c_k,a_k,\ell_k)\in\mathcal{K}$, compute
\[
(\sigma_k,\nu_k,r_k)=\mathcal{V}(q,a^\star,c_k,a_k),
\]
where $\sigma_k\in[0,1]$ is a confidence score, $\nu_k\in\{\textsf{pass},\textsf{fail}\}$ is the verdict under a default threshold $\tau=0.8$ (i.e., $\nu_k=\textsf{pass}$ iff $\sigma_k\ge\tau$), and $r_k$ is a brief issue report. We then form the verified subset
\[
\mathcal{K}_{\mathrm{ver}}=\{(c_k,a_k,\ell_k)\in\mathcal{K}:\nu_k=\textsf{pass}\},
\]
noting that $\ell_k$ is carried forward but not used by $\mathcal{V}$.

% \noindent 
% \paragraph{Answer matching \textit{\textsf{match}}.}
% We follow very weak answer matching approach. Given a generated answer $\hat{a}$ and gold $a^\star$, define $\textsf{match}:\mathcal{Y}\times\mathcal{Y}\to\{0,1\}$ by $\textsf{match}(\hat{a},a^\star)=1$ iff at least one holds: \textit{(i)} exact normalized equality, $\mathrm{norm}(\hat{a})=\mathrm{norm}(a^\star)$; \textit{(ii)} high token Jaccard, $J(P,G)=\frac{|P\cap G|}{\max(1,|P\cup G|)}\ge 0.75$, where $P=\mathrm{tokset}(\hat{a})$ and $G=\mathrm{tokset}(a^\star)$; \textit{(iii)} small-set containment, $\big(|P|\le 6 \land P\subseteq G\big)\ \lor\ \big(|G|\le 6 \land G\subseteq P\big)$; \textit{(iv)} high character similarity, $\mathrm{sim}(\mathrm{norm}(\hat{a}),\mathrm{norm}(a^\star))\ge\tau$, with $\tau=0.88$ and $\mathrm{sim}$ computed sequence matching algorithm\footnote{\url{https://docs.python.org/3/library/difflib.html}} Otherwise, $\textsf{match}(\hat{a},a^\star)=0$.

% To facilitate the answer matching, we use language-aware normalization $\mathrm{norm}(\cdot)$. For Arabic, we remove diacritics, and drop non-\{Arabic letters/digits/\_\} characters. For non-Arabic, we apply unicode normalization, lowercase, and remove non-\{a–z, 0–9\} characters. We set $\mathrm{tokset}(s)=\text{set}(\mathrm{norm}(s)\ \text{split on spaces})$.

\noindent
\paragraph{Answer matching \textit{\textsf{match}}.}
We follow very weak answer matching approach. Given a generated answer $\hat{a}$ and gold $a^\star$, define $\textsf{match}:\mathcal{Y}\times\mathcal{Y}\to\{0,1\}$ by $\textsf{match}(\hat{a},a^\star)=1$ iff at least one holds: \textit{(i)} exact normalized equality, $\mathrm{norm}(\hat{a})=\mathrm{norm}(a^\star)$; \textit{(ii)} high token Jaccard,
\[
J(P,G)=\frac{|P\cap G|}{\max(1,|P\cup G|)}\ge 0.75,
\]
where $P=\mathrm{tokset}(\hat{a})$ and $G=\mathrm{tokset}(a^\star)$; \textit{(iii)} small-set containment,
\[
\big(|P|\le 6 \land P\subseteq G\big)\ \lor\ \big(|G|\le 6 \land G\subseteq P\big);
\]
\textit{(iv)} high character similarity,
\[
\mathrm{sim}(\mathrm{norm}(\hat{a}),\mathrm{norm}(a^\star))\ge\tau,
\]
with $\tau=0.88$ and $\mathrm{sim}$ computed sequence matching algorithm\footnote{\url{https://docs.python.org/3/library/difflib.html}} otherwise, $\textsf{match}(\hat{a},a^\star)=0$.

To facilitate the answer matching, we use language-aware normalization $\mathrm{norm}(\cdot)$. For Arabic, we remove diacritics, and drop non-\{Arabic letters/digits/\_\} characters. For non-Arabic, we apply unicode normalization, lowercase, and remove non-\{a–z, 0–9\} characters. We set $\mathrm{tokset}(s)=\text{set}(\mathrm{norm}(s)\ \text{split on spaces})$.

\subsection{Manual Checking and Annotation}

\subsubsection{Preliminary Annotation}
We first conducted a targeted manual evaluation on small samples from each task, including dialectal translation and MCQ$\rightarrow$OEQ conversion. For each dialect, one native Arabic speaker who was also fluent in English reviewed the items. Annotators participated on a voluntary basis. Because this initial phase involved only one annotator per dialect, it served mainly as a lightweight quality check, but it still provided an early indication of data quality before larger-scale annotation.

To assess the generated items, we used a set of complementary rubrics covering both linguistic quality and task validity. Specifically, the rubrics evaluate: \textit{(i)} \emph{dialectal naturalness}, to measure whether the text sounds appropriate and idiomatic in the target dialect; \textit{(ii)} \emph{meaning preservation}, to check consistency with the source; \textit{(iii)} \emph{logical coherence}, to identify ill-formed, inconsistent, or contextually inappropriate items; \textit{(iv)} \emph{question-type appropriateness}, to ensure valid MCQ$\rightarrow$OEQ conversion; and \textit{(v)} \emph{linguistic quality and clarity}, to assess grammar, wording, and readability. The rubrics are defined as follows.
\begin{itemize} %[noitemsep,topsep=0pt,leftmargin=*,labelsep=.5em] 
    \item \textbf{Dialectal naturalness:} Do the question and, when applicable, the options sound fluent, idiomatic, and appropriate in the target dialect?
    \item \textbf{Meaning preservation:} Does the OEQ or translation convey the same meaning and intent as the original MCQ or source item?
    \item \textbf{Logical coherence:} Are the question and, when applicable, the options logically consistent, factually sound, and contextually appropriate?
    \item \textbf{Question-type appropriateness:} Is the format suitable for the content (i.e., MCQs are answerable by selection, while OEQs are genuinely open-ended)?
    \item \textbf{Linguistic quality and clarity:} Are grammar, wording, and orthography correct and easy for native speakers to understand?
\end{itemize}
We rate each dimension on a five-point Likert scale (from 1 to 5), which offers enough granularity to capture meaningful differences, includes a neutral midpoint for ambiguous cases, and supports simple aggregation across annotators and tasks. 
%Each dimension ranges from 1 to 5. %is scored on ${1,\dots,5}$.

Table~\ref{tab:manual_review} summarizes the initial manual annotation results for dialectal MCQs and MSA OEQs derived from MCQs. Overall scores are high, with an average of 4.4, indicating strong naturalness, meaning preservation, and linguistic quality. Maghrebi achieved the highest average score of 4.8, while English scored slightly lower at 4.1. The MSA MCQ$\rightarrow$OEQ transformation also performed well, with an average of 4.3, suggesting that the generated OEQs generally preserve the meaning and intent of the original MCQs.

% Table~\ref{tab:manual_review} summarizes the results of the initial manual annotation across dialectal MCQs and MSA OEQs derived from MCQs. Overall scores are high (average 4.4), indicating strong naturalness, meaning preservation, and linguistic quality. Maghrebi received the highest average (4.8), while English scored slightly lower (4.1). The MSA MCQ$\rightarrow$OEQ transformation also achieved strong scores (4.3), suggesting that the generated open-ended questions generally preserve the meaning and intent of the original MCQs. 

\begin{table}[htbp]
\centering
\small
\setlength{\tabcolsep}{3pt}
\scalebox{0.9}{%
\begin{tabular}{lcccccc}
\toprule
\textbf{Metric} & \textbf{MSA} & \textbf{Lv} & \textbf{Eg} & \textbf{Gf} & \textbf{En} & \textbf{Mg} \\
\midrule
Dialectal naturalness         & 4.2 & 4.4 & 4.3 & 4.3 & 4.2 & 4.7 \\
Meaning preservation           & 4.6 & 4.6 & 4.3 & 4.3 & 4.0 & 4.7 \\
Logical coherence              & 4.2 & 4.4 & 4.3 & 4.3 & 4.0 & 4.8 \\
Question-type appropriateness  & 4.1 & 4.7 & 4.3 & 4.4 & 4.2 & 4.8 \\
Linguistic quality and clarity & 4.2 & 4.5 & 4.4 & 4.4 & 4.0 & 4.8 \\ \midrule
\textbf{Average}               & \textbf{4.3} & \textbf{4.5} & \textbf{4.3} & \textbf{4.4} & \textbf{4.1} & \textbf{4.8} \\
\bottomrule
\end{tabular}
}
% \vspace{-0.2cm}
\caption{Average Likert score from manual annotations on a sample of 50 dialectal MCQs and MSA OEQs derived from MSA MCQs. Lv: Levantine, Eg: Egyptian, Gf: Gulf, En: English, Mg: Maghrebi.}
\label{tab:manual_review}
% \vspace{-0.25cm}
\end{table}

\subsubsection{Full Scale Annotation}
Following the preliminary annotation, we expanded our annotation task to the test set. Specifically, three independent annotators evaluated all MSA MCQ$\rightarrow$MSA OEQ conversions, assessing each converted \textbf{Question} across \emph{clarity}, \emph{naturalness in MSA}, \emph{being self-contained}, and \emph{appropriate scope}, and each corresponding \textbf{Answer} across \emph{correctness with respect to the original MCQ answer}, \emph{completeness}, \emph{conciseness}, and \emph{fluency}. Beyond Likert ratings, annotators also indicated whether revisions were needed and could provide optional comments when issues arose. This process helps verify that the generated OEQs remain consistent with the source MCQs while forming coherent and meaningful open-ended questions..
% }\subsubsection{Full-Scale Annotation}

For the \textbf{dialectal translation}, we performed full post-editing of the MSA MCQ$\rightarrow$Dialectal MCQs. Each dialectal MCQ test set was post-edited by a native speaker of the target dialect, who adapted the content to improve naturalness and linguistic authenticity while preserving the original meaning. This step ensures that the final dialectal MCQs reflect authentic usage patterns rather than literal translations. Detailed annotation guidelines and examples are provided in Appendix~\ref{sec:annotation_guidelines}.

\begin{table*}[!tbh]
\centering
\setlength{\tabcolsep}{5pt}
\scalebox{0.75}{
\begin{tabular}{lccccccc}
\toprule
\textbf{Model} & \textbf{MSA} & \textbf{Egyptian} & \textbf{Levantine} & \textbf{Magrebi} & \textbf{Gulf} & \textbf{English} & \textbf{Average} \\
\midrule
Falcon3-10B-Instruct & 46.05 & 43.95 & 44.10 & 42.70 & 45.15 & 66.50 & 48.48 \\
\rowcolor{ftblue} Falcon3-10B-Instruct FT & 57.65 & 55.15 & 54.25 & 53.60 & 55.95 & 71.90 & 58.17 \\
NileChat-3B & 67.55 & 64.75 & 64.65 & 64.45 & 66.00 & 65.15 & 65.00 \\
\rowcolor{ftblue} NileChat-3B FT & 69.20 & 67.75 & 67.65 & 66.90 & 67.45 & 69.05 & 67.76 \\ 
Fanar-1-9B-Instruct & 65.75 & 62.95 & 62.40 & 61.00 & 61.45 & 65.30 & 62.62 \\
\rowcolor{ftblue} Fanar-1-9B-Instruct FT & \textbf{72.55} & 69.85 & \textbf{70.55} & 69.70 & \textbf{70.75} & \textbf{72.65} & \textbf{70.70} \\
Qwen2.5-3B & 59.65 & 53.70 & 54.50 & 52.65 & 54.85 & 61.50 & 55.44 \\
\rowcolor{ftblue} Qwen2.5-3B FT & 63.75 & 62.80 & 62.80 & 62.45 & 62.60 & 69.55 & 64.04 \\
Qwen2.5-7B & 61.95 & 60.25 & 60.65 & 57.05 & 60.60 & 65.15 & 60.74 \\
\rowcolor{ftblue} Qwen2.5-7B FT & 67.50 & 65.85 & 65.95 & 63.25 & 66.00 & 71.50 & 66.51 \\
ALLaM-7B-Instruct-preview & 67.25 & 65.70 & 64.90 & 64.35 & 66.20 & 62.15 & 64.66 \\
\rowcolor{ftblue} ALLaM-7B-Instruct-preview FT & 71.95 & \textbf{70.55} & 69.85 & \textbf{69.85} & 70.40 & 67.70 & 69.67 \\
\midrule
\rowcolor{arabicavg} \textbf{Avg. Arabic-Centric} & 66.85 & 64.47 & 63.98 & 63.27 & 64.55 & 64.20 & 64.09 \\
\rowcolor{arabicavg} \textbf{Avg. Arabic-Centric FT} & 71.23 & 69.38 & 69.35 & 68.82 & 69.53 & 69.80 & 69.38 \\
\textbf{Avg. Base All} & 61.37 & 58.55 & 58.53 & 57.03 & 59.04 & 64.29 & 59.49 \\
\textbf{Avg. FT All} & 67.10 & 65.33 & 65.18 & 64.29 & 65.53 & 70.39 & 66.14 \\
\midrule
\rowcolor{gptgray} GPT-4.1 & 77.42 & 79.08 & 78.29 & 80.24 & 79.33 & 78.57 & 79.10 \\
\rowcolor{gptgray} GPT-5 & 79.59 & 79.10 & 78.88 & 77.70 & 79.31 & 77.17 & 78.43 \\
\bottomrule
\end{tabular}
}
% \vspace{-0.2cm}
\caption{MCQ accuracy (\%) across different language variants. Fine-tuned models (FT) are shaded in light blue, GPT models in gray, and averages for Arabic-centric models (NileChat, Fanar, ALLaM) are highlighted in light green. Bold values indicate the best-performing open model per dialect/language.}
\label{tab:mcq_results}
% \vspace{-0.3cm}
\end{table*}

\section{Experiments}
\label{sec:experiments}

\subsection{Models.} 
For the experiments, we used a range of open and closed-source multilingual and Arabic-centric models, covering capacities from small open frontier to frontier models. The models include \textit{Falcon3-10B-Instruct} \cite{malartic2024falcon2}, \textit{NileChat-3B} \cite{mekki2025nilechat}, \textit{Fanar-1-9B-Instruct} \cite{fanar2024}, \textit{Qwen2.5-3B} and \textit{Qwen2.5-7B} \cite{wang2024qwen2}, \textit{GPT-4.1} and \textit{GPT-5} \cite{openai2025gpt5}, and \textit{ALLaM-7B-Instruct-preview} \cite{bari2024allam}. This selection covers both high-performing proprietary and open models under 10B parameters, suitable for controlled fine-tuning and reproducible evaluation. 

\noindent
\subsection{Benchmarking.} 
All models were evaluated in a zero-shot setting across multiple language varieties. Prior work on cross-lingual prompting \cite{nativevsnonnative} has shown that non-native (English) prompts consistently outperform native prompts in reasoning and factual tasks, even for Arabic-centric models, while mixed prompts yield intermediate results. Following these findings, all evaluations in this study were conducted using English prompts. All prompts used are provided in Appendix~\ref{sec:other_prompts}.

All evaluations were conducted on the automatically generated dataset prior to the human post-editing and annotation stage.

\begin{table*}[!tbh]
\centering
\setlength{\tabcolsep}{4pt}
\scalebox{0.8}{
\begin{tabular}{lcccccccccccccc}
\toprule
\textbf{Model} & \multicolumn{2}{c}{\textbf{MSA}} & \multicolumn{2}{c}{\textbf{Egyptian}} & \multicolumn{2}{c}{\textbf{Levantine}} & \multicolumn{2}{c}{\textbf{Magrebi}} & \multicolumn{2}{c}{\textbf{Gulf}} & \multicolumn{2}{c}{\textbf{English}} & \multicolumn{2}{c}{\textbf{Average}} \\ \midrule
& F1 & RL & F1 & RL & F1 & RL & F1 & RL & F1 & RL & F1 & RL & F1 & RL \\
\midrule
Falcon3-10B-Instruct        & 0.43 & 0.12 & 0.41 & 0.10 & 0.41 & 0.09 & 0.41 & 0.09 & 0.41 & 0.10 & 0.54 & 0.23 & 0.44 & 0.12 \\
NileChat-3B                 & 0.48 & 0.17 & 0.49 & 0.18 & 0.50 & 0.17 & 0.50 & 0.18 & 0.49 & 0.17 & 0.49 & 0.15 & 0.49 & 0.17 \\
Fanar-1-9B-Instruct         & 0.52 & 0.20 & 0.50 & 0.17 & 0.50 & 0.16 & 0.50 & 0.17 & 0.51 & 0.17 & \textbf{0.53} & 0.18 & 0.51 & 0.18 \\
Qwen2.5-3B                  & 0.45 & 0.13 & 0.43 & 0.11 & 0.43 & 0.10 & 0.44 & 0.11 & 0.44 & 0.11 & 0.47 & 0.11 & 0.44 & 0.11 \\
Qwen2.5-7B                  & \textbf{0.55} & 0.24 & \textbf{0.51} & 0.20 & \textbf{0.51} & 0.19 & \textbf{0.53} & 0.21 & \textbf{0.52} & 0.20 & \textbf{0.53} & 0.20 & \textbf{0.53} & 0.21 \\
ALLaM-7B-Instruct           & 0.49 & 0.20 & 0.47 & 0.16 & 0.47 & 0.15 & 0.46 & 0.15 & 0.48 & 0.17 & 0.52 & 0.22 & 0.48 & 0.17 \\ \midrule
\rowcolor{arabicavg} \textbf{Avg. Arabic-Centric}       & 0.50 & 0.19 & 0.49 & 0.17 & 0.49 & 0.16 & 0.49 & 0.17 & 0.49 & 0.17 & 0.51 & 0.18 & 0.50 & 0.17 \\
\textbf{Avg. Base}       & 0.49 & 0.18 & 0.47 & 0.16 & 0.47 & 0.14 & 0.47 & 0.15 & 0.47 & 0.15 & 0.51 & 0.18 & 0.48 & 0.16 \\
\midrule
\rowcolor{gptgray} GPT-4.1  & 0.55 & 0.27 & 0.53 & 0.24 & 0.53 & 0.21 & 0.54 & 0.24 & 0.54 & 0.24 & 0.56 & 0.25 & 0.54 & 0.24 \\
\rowcolor{gptgray} GPT-5    & 0.57 & 0.28 & 0.54 & 0.24 & 0.54 & 0.22 & 0.55 & 0.24 & 0.55 & 0.25 & 0.54 & 0.22 & 0.55 & 0.24 \\
\bottomrule
\end{tabular}
}
% \vspace{-0.2cm}
\caption{OEQ performance across different language variants. F1: BERTScore F1, RL: Rouge-L. Averages for Arabic-centric models (NileChat, Fanar, ALLaM) are highlighted in light green. GPT models are shaded in gray.}
\label{tab:oeq_results}
% \vspace{-0.35cm}
\end{table*}

\noindent
\subsection{Training.} 
We adopt fine-tuning configurations consistent with prior work on Arabic cultural QA tasks, as reported in \cite{cultran2025palmx}. Fine-tuning is conducted over 3 epochs using LoRA adapters \cite{hu2022lora}, with a maximum sequence length of 512 for MCQ training and 2048 for OEQ training. The learning rate is set to $2\times10^{-4}$, with a LoRA rank of 64 and $\alpha=16$. All models are fine-tuned for MCQ evaluation, while only \texttt{ALLaM-7B-Instruct-preview} is fine-tuned for the OEQ task.

\noindent
\subsection{Evaluation and Metrics.}
For MCQ, we report accuracy, which is an standard metric for MCQ. For OEQ, we employ semantic evaluation using \textsc{BERTScore} \cite{zhang2020bertscore} and \textsc{ROUGE-L} \cite{lin2004rouge} to assess precision, recall, and overall semantic overlap with the gold answers. Arabic responses are evaluated using \texttt{arabert-v2} \cite{antoun-etal-2020-arabert}, and English responses with \texttt{bert-base-uncased}. This setup allows direct comparability between multilingual and dialectal outputs across all evaluated models. Additionally, for OEQ, we use GPT-4.1 as LLM-as-judge following MT-Bench~\citep{bai-etal-2024-mt}, where responses are rated on a 1 to 10 rubric (helpfulness, relevance, accuracy, faithfulness).

\section{Results}

We compare performance across four conditions: \textit{(i)} MCQ base vs. fine-tuned, \textit{(ii)} OEQ base, \textit{(iii)} OEQ fine-tuned without CoT, and \textit{(iv)} OEQ fine-tuned with CoT. Tables~\ref{tab:mcq_results}, \ref{tab:oeq_results}, and \ref{tab:oeq_with_llm_judge} present the results for the MCQ, OEQ, and OEQ (with vs. without CoT) evaluations, respectively.

\begin{table}[!tbh]
\centering
\setlength{\tabcolsep}{4pt}
\scalebox{0.7}{
\begin{tabular}{l|ccc|ccc|ccc}
\toprule
\textbf{Lang} & \multicolumn{3}{c|}{\textbf{Base}} & \multicolumn{3}{c|}{\textbf{FT}} & \multicolumn{3}{c}{\textbf{FT with COT}} \\
\midrule
 & \textbf{J} & \textbf{F1} & \textbf{RL} & \textbf{J} & \textbf{F1} & \textbf{RL} & \textbf{J} & \textbf{F1} & \textbf{RL} \\
\midrule
MSA       & 5.50 & 0.49 & 0.20 & 6.02 & 0.76 & 0.56 & 6.12 & 0.70 & 0.48 \\ 
Eg  & 4.93 & 0.47 & 0.16 & 5.90 & 0.71 & 0.46 & 6.10 & 0.66 & 0.41 \\
Lv & 4.95 & 0.47 & 0.15 & 5.93 & 0.70 & 0.45 & 6.13 & 0.66 & 0.40 \\
Mg   & 4.80 & 0.46 & 0.15 & 5.88 & 0.70 & 0.45 & 6.08 & 0.65 & 0.39 \\
Gf      & 4.97 & 0.48 & 0.17 & 5.94 & 0.70 & 0.45 & 6.14 & 0.66 & 0.41 \\
En   & 4.49 & 0.52 & 0.22 & 5.55 & 0.74 & 0.57 & 5.48 & 0.67 & 0.43 \\
\midrule
\textbf{Avg.} & 4.94 & 0.48 & 0.17 & 5.87 & 0.72 & 0.49 & 6.01 & 0.67 & 0.42 \\
\bottomrule
\end{tabular}
}
% \vspace{-0.2cm}
\caption{Performance on the OEQ across ALLaM-7B base, fine-tuned (FT), and fine-tuned with CoT models. J: LLM-as-a-judge.}
\label{tab:oeq_with_llm_judge}
% \vspace{-0.4cm}
\end{table}

\noindent
\paragraph{Performance Gap for MCQ.} 
As presented in Table~\ref{tab:mcq_results}, the average performance among the Arabic language variants is relatively higher for MSA across open models, followed by Gulf, Egyptian, and others. The average performance for English is higher compared to Arabic across open models, mainly due to the strong performance of non-Arabic-centric models such as Falcon and Qwen. The average performance for Arabic-centric models in the base and fine-tuned (FT) settings is 64.09\% and 69.38\%, respectively. 

The performance of closed models (i.e., GPT*) are higher than closed models in all language variants. The MCQ performance for MSA is highly comparable with the PalmX shared task results where top-system achieved an accuracy of 72.15\% \cite{alwajih2025palmx}.

Among the smaller open models (i.e., size 3B), in the base setting, \texttt{NileChat-3B} achieves the highest average accuracy of 65.43, while \texttt{Fanar-1-9B-Instruct} is the best-performing fine-tuned model with an accuracy of 71.01. Among the open models, the fine-tuned \texttt{ALLaM-7B-Instruct} performs best for Egyptian and Maghrebi, whereas \texttt{Fanar-1-9B-Instruct-FT} achieves the highest performance for MSA, Le, Gf, and En.

\noindent
\paragraph{Performance Gap for OEQ.}
Across language variants, we observe a pattern consistent with MCQ results: the average F1 for MSA exceeds that of other Arabic dialects; however, the gap is smaller than in the MCQ setting (Table~\ref{tab:oeq_results}). Similarly, English achieves higher F1 than the Arabic variants. Notably, for OEQ, the \texttt{Qwen2.5-7B} open model outperforms the other open models, including Arabic-centric ones.

Among all base models, \texttt{GPT-5} achieves the highest overall performance (\(F1=0.55\)), followed closely by \texttt{GPT-4.1} (\(F1=0.54\)). GPT-5 performs best on MSA, while GPT-4.1 shows strong results on both English and MSA.

% Across different language variants, we observe a consistent for performance difference like MCQ -- average F1 for MSA is higher compared to other dialects, however, the gap is minimal than MCQ, as presented in \ref{tab:oeq_results}. Similarly, the F1 for English is higher than Arabic variants. Surprising, for OEQ, the Qwen (7B) model is outperforming all other models including Arabic-centric ones. 

% Among all base models, \textbf{GPT-5} achieves the highest overall performance (F1 = 0.55), followed closely by \textbf{GPT-4.1} (F1 = 0.54). GPT-5 performs best on MSA, while GPT-4.1 shows strong results on both English and MSA. 
 
% Among all base models, \textbf{GPT-5} achieves the highest overall performance (F1 = 0.55, Rouge-L = 0.24), followed closely by \textbf{GPT-4.1} (F1 = 0.54, Rouge-L = 0.24).  GPT-5 performs best on MSA, while GPT-4.1 shows strong results on both English and MSA. 
% Among open models, Qwen 7B leads with average F1 0.53 and rogueL1 of 0.21 with \textbf{Fanar-1-9B-Instruct (FT)} coming second with 71.01\% average. Across dialects, English annd MSA performance was overall the highest, while all ohters dialects were just slightly lower.

\noindent
\paragraph{Did CoT help for OEQ?}
In Table \ref{tab:oeq_with_llm_judge}, we report the performance of OEQ with a comparison to the base model, fine-tuning without CoT (FT), and fine-tuning with CoT. Other than F1 and Rouge-L score, we also report LLM-as-a-judge scores. On token-overlap metrics, FT yields the strongest scores (F1/RL), whereas the CoT-tuned model achieves the highest average \emph{LLM-as-a-judge} score. This difference indicates that CoT improves \emph{semantic acceptability} but reduces \emph{lexical overlap} with the references. A manual pass over low-F1 cases shows that the CoT model frequently returns briefer answers that judges deem correct, yet they share fewer n-grams with the (often longer) gold strings, decreasing F1 and RL. Overall, CoT helps on judged correctness but not on n-gram overlap. 

This pattern aligns with prior findings that CoT is not uniformly beneficial. For instance, \citet{zhu-etal-2025-rationales} show that adding rationales can sometimes hurt performance, while \citet{li2025small} find that fine-tuning smaller models on lengthy, teacher-generated CoT traces performs no better, or worse, than training without CoT. Together with our results, these observations highlight the need to examine when CoT is advantageous, particularly regarding task type, rationale length, and model size.

% This pattern aligns with prior findings that CoT is not uniformly beneficial. For example, \citet{zhu-etal-2025-rationales} report that adding rationales across many tasks can fail to help and may even hurt performance, and \citet{li2025small} observe that fine-tuning smaller models on long, teacher-generated CoT traces performs no better—or worse—than training without CoT. Taken together with our results, these observations motivate further study of when CoT is advantageous, including the roles of task, rationale length, and model size in FT.

\section{Conclusions and Future Work}
\label{sec:conclusions}
% \textcolor{blue}{
We presented a comprehensive pipeline for converting MCQ into OEQ and extended an existing MCQ dataset across multiple language varieties, including MSA, English, and several Arabic dialects. To our knowledge, \textbf{\textit{ArabicCulturalQA}} is the \textit{\textbf{first}} Arabic cultural \textbf{OEQ resource with parallel dialectal variants} alongside English, providing a foundation for culturally grounded evaluation beyond MSA. Though this dataset is based on PalmX, however, we have extensively extended it: QAs are \emph{parallelly aligned} across all language variants. 
% In addition, we conduct systematic human validation and
We have manually checked MCQ to OEQ mapping and post-edited the translated dialectal MCQs
% , including full annotation of the MSA OEQ test set and post-editing of dialectal MCQs 
by native speakers to ensure naturalness and fidelity. We benchmarked the dataset using a set of open, closed, and fine-tuned models. Fine-tuned models consistently outperform their base counterparts yet still lag behind strong closed models. Performance is generally higher for MSA than for dialects. Arabic-centric models show advantages on Arabic variants for MCQ but smaller gains on OEQ, highlighting the added difficulty of generative, culturally grounded answering. Our initial CoT results improve judged correctness but yield mixed n-gram–based scores. Future work includes annotation and post-editing of the remaining dialectal OEQ test set, variety-aware normalisation and scoring, and extending to additional low-resource languages and modalities.
% }

\section{Ethics statement}
% \textcolor{blue}{
We do not anticipate ethical concerns arising from this work. We build on publicly available datasets that permit research use, and we comply with their licenses and terms. For the manual annotations, contributors participated %voluntarily 
after being fully briefed on the task and its purpose. Initial annotators participated on a voluntary basis, while annotators involved in the expanded annotation and post-editing stages were compensated for their work. No personal or sensitive data were collected beyond what is contained in the source datasets.   
% }

\section{Limitations}
\label{sec:limitations}
% \textcolor{blue}
% {
Our extensions to publicly available Arabic-dialect datasets rely on LLM-assisted translation and MCQ$\to$OEQ conversion, which may introduce modeling biases (e.g., paraphrase drift, dialectal normalization) and occasional errors. Although we subsequently performed systematic annotation and dialectal post-editing to improve naturalness and quality, the experimental results reported in this paper were computed on the automatically generated versions of the datasets prior to human post-editing. As a result, the reported benchmarks may underestimate the quality and usability of the finalized released datasets.
% }

\section*{References}
\bibliographystyle{lrec2026-natbib}
\bibliography{bib/bibliography}

@article{alam2025everydaymmqa,
  title = {{EverydayMMQA}: A Multilingual and Multimodal Framework for Culturally Grounded Spoken Visual QA},
  author = {Alam, Firoj and Shahroor, Ali Ezzat and Hasan, Md. Arid and Ali, Zien Sheikh and Bhatti, Hunzalah Hassan and Kmainasi, Mohamed Bayan and Chowdhury, Shammur Absar and Mousi, Basel and Dalvi, Fahim and Durrani, Nadir and Milic-Frayling, Natasa},
  journal = {arXiv preprint arXiv:2510.06371},
  year = {2025},
}

@article{li2025small,
  title={Small models struggle to learn from strong reasoners},
  author={Li, Yuetai and Yue, Xiang and Xu, Zhangchen and Jiang, Fengqing and Niu, Luyao and Lin, Bill Yuchen and Ramasubramanian, Bhaskar and Poovendran, Radha},
  journal={arXiv preprint arXiv:2502.12143},
  year={2025}
}

@inproceedings{zhu-etal-2025-rationales,
    title = "Rationales Are Not Silver Bullets: Measuring the Impact of Rationales on Model Performance and Reliability",
    author = "Zhu, Chiwei  and
      Xu, Benfeng  and
      Yang, An  and
      Lin, Junyang  and
      Wang, Quan  and
      Zhou, Chang  and
      Mao, Zhendong",
    editor = "Che, Wanxiang  and
      Nabende, Joyce  and
      Shutova, Ekaterina  and
      Pilehvar, Mohammad Taher",
    booktitle = "Findings of the Association for Computational Linguistics: ACL 2025",
    month = jul,
    year = "2025",
    address = "Vienna, Austria",
    publisher = "Association for Computational Linguistics",
    url = "https://aclanthology.org/2025.findings-acl.302/",
    doi = "10.18653/v1/2025.findings-acl.302",
    pages = "5808--5835",
    ISBN = "979-8-89176-256-5"
}

@inproceedings{bai-etal-2024-mt,
    title = "{MT}-Bench-101: A Fine-Grained Benchmark for Evaluating Large Language Models in Multi-Turn Dialogues",
    author = "Bai, Ge  and
      Liu, Jie  and
      Bu, Xingyuan  and
      He, Yancheng  and
      Liu, Jiaheng  and
      Zhou, Zhanhui  and
      Lin, Zhuoran  and
      Su, Wenbo  and
      Ge, Tiezheng  and
      Zheng, Bo  and
      Ouyang, Wanli",
    editor = "Ku, Lun-Wei  and
      Martins, Andre  and
      Srikumar, Vivek",
    booktitle = "Proceedings of the 62nd Annual Meeting of the Association for Computational Linguistics (Volume 1: Long Papers)",
    month = aug,
    year = "2024",
    address = "Bangkok, Thailand",
    publisher = "Association for Computational Linguistics",
    url = "https://aclanthology.org/2024.acl-long.401/",
    doi = "10.18653/v1/2024.acl-long.401",
    pages = "7421--7454"
}

@article{zelikman2022star,
  title={{STaR}: Bootstrapping reasoning with reasoning},
  author={Zelikman, Eric and Wu, Yuhuai and Mu, Jesse and Goodman, Noah},
  journal={Advances in Neural Information Processing Systems},
  volume={35},
  pages={15476--15488},
  year={2022}
}

@article{chandak2025answer,
  title={Answer Matching Outperforms Multiple Choice for Language Model Evaluation},
  author={Chandak, Nikhil and Goel, Shashwat and Prabhu, Ameya and Hardt, Moritz and Geiping, Jonas},
  journal={arXiv preprint arXiv:2507.02856},
  year={2025}
}

@article{malartic2024falcon2,
  title={Falcon2-11b technical report},
  author={Malartic, Quentin and Chowdhury, Nilabhra Roy and Cojocaru, Ruxandra and Farooq, Mugariya and Campesan, Giulia and Djilali, Yasser Abdelaziz Dahou and Narayan, Sanath and Singh, Ankit and Velikanov, Maksim and Boussaha, Basma El Amel and others},
  journal={arXiv preprint arXiv:2407.14885},
  year={2024}
}

@inproceedings{puerto-etal-2025-fine,
    title = "Fine-Tuning on Diverse Reasoning Chains Drives Within-Inference {C}o{T} Refinement in {LLM}s",
    author = "Puerto, Haritz  and
      Chubakov, Tilek  and
      Zhu, Xiaodan  and
      Tayyar Madabushi, Harish  and
      Gurevych, Iryna",
    editor = "Che, Wanxiang  and
      Nabende, Joyce  and
      Shutova, Ekaterina  and
      Pilehvar, Mohammad Taher",
    booktitle = "Proceedings of the 63rd Annual Meeting of the Association for Computational Linguistics (Volume 1: Long Papers)",
    month = jul,
    year = "2025",
    address = "Vienna, Austria",
    publisher = "Association for Computational Linguistics",
    url = "https://aclanthology.org/2025.acl-long.191/",
    doi = "10.18653/v1/2025.acl-long.191",
    pages = "3789--3808",
    ISBN = "979-8-89176-251-0"
}

@inproceedings{qin-etal-2023-cross,
    title = "Cross-lingual Prompting: Improving Zero-shot Chain-of-Thought Reasoning across Languages",
    author = "Qin, Libo  and
      Chen, Qiguang  and
      Wei, Fuxuan  and
      Huang, Shijue  and
      Che, Wanxiang",
    editor = "Bouamor, Houda  and
      Pino, Juan  and
      Bali, Kalika",
    booktitle = "Proceedings of the 2023 Conference on Empirical Methods in Natural Language Processing",
    month = dec,
    year = "2023",
    address = "Singapore",
    publisher = "Association for Computational Linguistics",
    url = "https://aclanthology.org/2023.emnlp-main.163/",
    doi = "10.18653/v1/2023.emnlp-main.163",
    pages = "2695--2709"
}

@article{molfese2025right,
  title={Right Answer, Wrong Score: Uncovering the Inconsistencies of LLM Evaluation in Multiple-Choice Question Answering},
  author={Molfese, Francesco Maria and Moroni, Luca and Gioffr{\'e}, Luca and Scir{\`e}, Alessandro and Conia, Simone and Navigli, Roberto},
  journal={arXiv preprint arXiv:2503.14996},
  year={2025}
}

@article{kojima2022large,
  title={Large language models are zero-shot reasoners},
  author={Kojima, Takeshi and Gu, Shixiang Shane and Reid, Machel and Matsuo, Yutaka and Iwasawa, Yusuke},
  journal={Advances in neural information processing systems},
  volume={35},
  pages={22199--22213},
  year={2022}
}

@inproceedings{li2024can,
  title={Can Multiple-choice Questions Really Be Useful in Detecting the Abilities of LLMs?},
  author={Li, Wangyue and Li, Liangzhi and Xiang, Tong and Liu, Xiao and Deng, Wei and Garcia, Noa},
  booktitle={Proceedings of the 2024 Joint International Conference on Computational Linguistics, Language Resources and Evaluation (LREC-COLING 2024)},
  pages={2819--2834},
  year={2024}
}

@article{raman2025reasoning,
  title={Reasoning models are test exploiters: Rethinking multiple-choice},
  author={Raman, Narun and Lundy, Taylor and Leyton-Brown, Kevin},
  journal={arXiv preprint arXiv:2507.15337},
  year={2025}
}

@inproceedings{alwajih-etal-2025-palm,
    title = "Palm: A Culturally Inclusive and Linguistically Diverse Dataset for {A}rabic {LLM}s",
    author = "Alwajih, Fakhraddin  and
      El Mekki, Abdellah  and
      Magdy, Samar Mohamed  and
      Elmadany, AbdelRahim A.  and
      Nacar, Omer  and
      Nagoudi, El Moatez Billah  and
      Abdel-Salam, Reem  and
      Atwany, Hanin  and
      Nafea, Youssef  and
      Yahya, Abdulfattah Mohammed  and
      Alhamouri, Rahaf  and
      Alsayadi, Hamzah A.  and
      Zayed, Hiba  and
      Shatnawi, Sara  and
      Sibaee, Serry  and
      Ech-chammakhy, Yasir  and
      Al-Dhabyani, Walid  and
      Ali, Marwa Mohamed  and
      Jarraya, Imen  and
      El-Shangiti, Ahmed Oumar  and
      Alraeesi, Aisha  and
      AL-Ghrawi, Mohammed Anwar  and
      Al-Batati, Abdulrahman S.  and
      Mohamed, Elgizouli  and
      Elgindi, Noha Taha  and
      Saeed, Muhammed  and
      Atou, Houdaifa  and
      Yahia, Issam Ait  and
      Bouayad, Abdelhak  and
      Machrouh, Mohammed  and
      Makouar, Amal  and
      Alkawi, Dania  and
      Mohamed, Mukhtar  and
      Abdelfadil, Safaa Taher  and
      Ounnoughene, Amine Ziad  and
      Rouabhia, Anfel  and
      Assi, Rwaa  and
      Sorkatti, Ahmed  and
      Tourad, Mohamedou Cheikh  and
      Koubaa, Anis  and
      Berrada, Ismail  and
      Jarrar, Mustafa  and
      Shehata, Shady  and
      Abdul-Mageed, Muhammad",
    editor = "Che, Wanxiang  and
      Nabende, Joyce  and
      Shutova, Ekaterina  and
      Pilehvar, Mohammad Taher",
    booktitle = "Proceedings of the 63rd Annual Meeting of the Association for Computational Linguistics (Volume 1: Long Papers)",
    month = jul,
    year = "2025",
    address = "Vienna, Austria",
    publisher = "Association for Computational Linguistics",
    url = "https://aclanthology.org/2025.acl-long.1579/",
    doi = "10.18653/v1/2025.acl-long.1579",
    pages = "32871--32894",
    ISBN = "979-8-89176-251-0"
}

@inproceedings{muennighoff-etal-2023-crosslingual,
    title = "Crosslingual Generalization through Multitask Finetuning",
    author = "Muennighoff, Niklas  and
      Wang, Thomas  and
      Sutawika, Lintang  and
      Roberts, Adam  and
      Biderman, Stella  and
      Le Scao, Teven  and
      Bari, M Saiful  and
      Shen, Sheng  and
      Yong, Zheng Xin  and
      Schoelkopf, Hailey  and
      Tang, Xiangru  and
      Radev, Dragomir  and
      Aji, Alham Fikri  and
      Almubarak, Khalid  and
      Albanie, Samuel  and
      Alyafeai, Zaid  and
      Webson, Albert  and
      Raff, Edward  and
      Raffel, Colin",
    editor = "Rogers, Anna  and
      Boyd-Graber, Jordan  and
      Okazaki, Naoaki",
    booktitle = "Proceedings of the 61st Annual Meeting of the Association for Computational Linguistics (Volume 1: Long Papers)",
    month = jul,
    year = "2023",
    address = "Toronto, Canada",
    publisher = "Association for Computational Linguistics",
    url = "https://aclanthology.org/2023.acl-long.891/",
    doi = "10.18653/v1/2023.acl-long.891",
    pages = "15991--16111"
}

@InProceedings{nativevsnonnative,
author="Kmainasi, Mohamed Bayan
and Khan, Rakif
and Shahroor, Ali Ezzat
and Bendou, Boushra
and Hasanain, Maram
and Alam, Firoj",
editor="Barhamgi, Mahmoud
and Wang, Hua
and Wang, Xin",
title="Native vs Non-native Language Prompting: A Comparative Analysis",
booktitle="Web Information Systems Engineering -- WISE 2024",
year="2025",
publisher="Springer Nature Singapore",
address="Singapore",
pages="406--420",
isbn="978-981-96-0576-7"
}

@misc{openai2025gpt5,
  title     = {GPT-5 Technical Overview},
  author    = {OpenAI},
  year      = {2025},
  howpublished = {\url{https://openai.com/research/gpt-5}}
}

@misc{cultran2025palmx,
      title={CultranAI at PalmX 2025: Data Augmentation for Cultural Knowledge Representation}, 
      author={Hunzalah Hassan Bhatti and Youssef Ahmed and Md Arid Hasan and Firoj Alam},
      year={2025},
      eprint={2508.17324},
      archivePrefix={arXiv},
      primaryClass={cs.CL},
      url={https://arxiv.org/abs/2508.17324}, 
}

@inproceedings{zhang2020bertscore,
  title     = {{BERTScore}: Evaluating Text Generation with BERT},
  author    = {Tianyi Zhang and Varsha Kishore and Felix Wu and Kilian Q. Weinberger and Yoav Artzi},
  booktitle = {Proceedings of ICLR 2020},
  year      = {2020}
}

@inproceedings{lin2004rouge,
  title     = {{ROUGE}: A Package for Automatic Evaluation of Summaries},
  author    = {Chin-Yew Lin},
  booktitle = {Text Summarization Branches Out: Proceedings of the ACL Workshop},
  year      = {2004}
}

@article{myrzakhan2024open,
  title={Open-llm-leaderboard: From multi-choice to open-style questions for llms evaluation, benchmark, and arena},
  author={Myrzakhan, Aidar and Bsharat, Sondos Mahmoud and Shen, Zhiqiang},
  journal={arXiv preprint arXiv:2406.07545},
  year={2024}
}

@inproceedings{sadallah-etal-2025-commonsense,
    title = "Commonsense Reasoning in {A}rab Culture",
    author = "Sadallah, Abdelrahman  and
      Tonga, Junior Cedric  and
      Almubarak, Khalid  and
      Almheiri, Saeed  and
      Atif, Farah  and
      Qwaider, Chatrine  and
      Kadaoui, Karima  and
      Shatnawi, Sara  and
      Alesh, Yaser  and
      Koto, Fajri",
    editor = "Che, Wanxiang  and
      Nabende, Joyce  and
      Shutova, Ekaterina  and
      Pilehvar, Mohammad Taher",
    booktitle = "Proceedings of the 63rd Annual Meeting of the Association for Computational Linguistics (Volume 1: Long Papers)",
    month = jul,
    year = "2025",
    address = "Vienna, Austria",
    publisher = "Association for Computational Linguistics",
    url = "https://aclanthology.org/2025.acl-long.380/",
    doi = "10.18653/v1/2025.acl-long.380",
    pages = "7695--7710",
    ISBN = "979-8-89176-251-0"
}

@article{yu2025cotselfinstruct,
  title        = {CoT-Self-Instruct: Building High-Quality Synthetic Data for Reasoning and Non-Reasoning Tasks},
  author       = {Ping Yu and Jack Lanchantin and Tianlu Wang and Weizhe Yuan and Olga Golovneva and Ilia Kulikov and Sainbayar Sukhbaatar and Jason Weston and Jing Xu},
  journal      = {arXiv preprint arXiv:2507.23751},
  year         = {2025},
  url          = {https://arxiv.org/abs/2507.23751}
}

@inproceedings{alwajih2025palmx,
  title     = {{PalmX 2025}: The First Shared Task on Benchmarking LLMs on Arabic and Islamic Culture},
  author    = {Alwajih, Fakhraddin and El Mekki, Abdellah and Mubarak, Hamdy and Hawasly, Majd and Mohamed, Abubakr and Abdul-Mageed, Muhammad},
  booktitle = {Proceedings of the Third Arabic Natural Language Processing Conference (ArabicNLP 2025)},
  month     = nov,
  year      = {2025},
  address   = {Suzhou, China},
  publisher = {Association for Computational Linguistics},
  note      = {Co-located with EMNLP 2025, November 5--9}
}

@article{mekki2025nilechat,
  title={NileChat: Towards Linguistically Diverse and Culturally Aware LLMs for Local Communities},
  author={Mekki, Abdellah El and Atou, Houdaifa and Nacar, Omer and Shehata, Shady and Abdul-Mageed, Muhammad},
  journal={arXiv preprint arXiv:2505.18383},
  year={2025}
}

@article{wang2024qwen2,
  title={Qwen2-vl: Enhancing vision-language model's perception of the world at any resolution},
  author={Wang, Peng and Bai, Shuai and Tan, Sinan and Wang, Shijie and Fan, Zhihao and Bai, Jinze and Chen, Keqin and Liu, Xuejing and Wang, Jialin and Ge, Wenbin and others},
  journal={arXiv preprint arXiv:2409.12191},
  year={2024}
}

@article{alam2025nativqaframeworkenablingllms,
  title = {{NativQA Framework}: Enabling LLMs with Native, Local, and Everyday Knowledge},
  author = {Alam, Firoj and Hasan, Md Arid and Laskar, Sahinur Rahman and Kutlu, Mucahid and Chowdhury, Shammur Absar},
  journal = {arXiv preprint arXiv:2504.05995},
  year = {2025},
  url = {https://arxiv.org/abs/2504.05995},
}

@article{hu2022lora,
  title={Lora: Low-rank adaptation of large language models.},
  author={Hu, Edward J and Shen, Yelong and Wallis, Phillip and Allen-Zhu, Zeyuan and Li, Yuanzhi and Wang, Shean and Wang, Lu and Chen, Weizhu and others},
  journal={ICLR},
  volume={1},
  number={2},
  pages={3},
  year={2022}
}

@inproceedings{bari2024allam,
    title={{ALL}aM: Large Language Models for Arabic and English},
    author={M Saiful Bari and Yazeed Alnumay and Norah A. Alzahrani and Nouf M. Alotaibi and Hisham Abdullah Alyahya and Sultan AlRashed and Faisal Abdulrahman Mirza and Shaykhah Z. Alsubaie and Hassan A. Alahmed and Ghadah Alabduljabbar and Raghad Alkhathran and Yousef Almushayqih and Raneem Alnajim and Salman Alsubaihi and Maryam Al Mansour and Saad Amin Hassan and Dr. Majed Alrubaian and Ali Alammari and Zaki Alawami and Abdulmohsen Al-Thubaity and Ahmed Abdelali and Jeril Kuriakose and Abdalghani Abujabal and Nora Al-Twairesh and Areeb Alowisheq and Haidar Khan},
    booktitle={The Thirteenth International Conference on Learning Representations},
    year={2025},
    url={https://openreview.net/forum?id=MscdsFVZrN}
}

@article{wang2023aligning,
  title={Aligning large language models with human: A survey},
  author={Wang, Yufei and Zhong, Wanjun and Li, Liangyou and Mi, Fei and Zeng, Xingshan and Huang, Wenyong and Shang, Lifeng and Jiang, Xin and Liu, Qun},
  journal={arXiv preprint arXiv:2307.12966},
  year={2023}
}

@inproceedings{abdelali-etal-2024-larabench,
    title = "{LA}ra{B}ench: Benchmarking {A}rabic {AI} with Large Language Models",
    author = "Abdelali, Ahmed  and
      Mubarak, Hamdy  and
      Chowdhury, Shammur  and
      Hasanain, Maram  and
      Mousi, Basel  and
      Boughorbel, Sabri  and
      Abdaljalil, Samir  and
      El Kheir, Yassine  and
      Izham, Daniel  and
      Dalvi, Fahim  and
      Hawasly, Majd  and
      Nazar, Nizi  and
      Elshahawy, Youssef  and
      Ali, Ahmed  and
      Durrani, Nadir  and
      Milic-Frayling, Natasa  and
      Alam, Firoj",
    editor = "Graham, Yvette  and
      Purver, Matthew",
    booktitle = "Proceedings of the 18th Conference of the European Chapter of the Association for Computational Linguistics (Volume 1: Long Papers)",
    month = mar,
    year = "2024",
    address = "St. Julian{'}s, Malta",
    publisher = "Association for Computational Linguistics",
    url = "https://aclanthology.org/2024.eacl-long.30/",
    pages = "487--520"
}

@inproceedings{mousi-etal-2025-aradice,
    title = "{A}ra{D}i{CE}: Benchmarks for Dialectal and Cultural Capabilities in {LLM}s",
    author = "Mousi, Basel  and
      Durrani, Nadir  and
      Ahmad, Fatema  and
      Hasan, Md Arid  and
      Hasanain, Maram  and
      Kabbani, Tameem  and
      Dalvi, Fahim  and
      Chowdhury, Shammur Absar  and
      Alam, Firoj",
    editor = "Rambow, Owen  and
      Wanner, Leo  and
      Apidianaki, Marianna  and
      Al-Khalifa, Hend  and
      Eugenio, Barbara Di  and
      Schockaert, Steven",
    booktitle = "Proceedings of the 31st International Conference on Computational Linguistics",
    month = jan,
    year = "2025",
    address = "Abu Dhabi, UAE",
    publisher = "Association for Computational Linguistics",
    url = "https://aclanthology.org/2025.coling-main.283/",
    pages = "4186--4218"
}

@article{fanar2024,
      title={Fanar: An Arabic-Centric Multimodal Generative AI Platform}, 
      author={Fanar Team and Ummar Abbas and Mohammad Shahmeer Ahmad and Firoj Alam and Enes Altinisik and Ehsannedin Asgari and Yazan Boshmaf and Sabri Boughorbel and Sanjay Chawla and Shammur Chowdhury and Fahim Dalvi and Kareem Darwish and Nadir Durrani and Mohamed Elfeky and Ahmed Elmagarmid and Mohamed Eltabakh and Masoomali Fatehkia and Anastasios Fragkopoulos and Maram Hasanain and Majd Hawasly and Mus'ab Husaini and Soon-Gyo Jung and Ji Kim Lucas and Walid Magdy and Safa Messaoud and Abubakr Mohamed and Tasnim Mohiuddin and Basel Mousi and Hamdy Mubarak and Ahmad Musleh and Zan Naeem and Mourad Ouzzani and Dorde Popovic and Amin Sadeghi and Husrev Taha Sencar and Mohammed Shinoy and Omar Sinan and Yifan Zhang and Ahmed Ali and Yassine El Kheir and Xiaosong Ma and Chaoyi Ruan},
      year={2025},
      eprint={2501.13944},
      archivePrefix={arXiv},
      primaryClass={cs.CL},
      url={https://arxiv.org/abs/2501.13944}, 
}

@inproceedings{hasan-etal-2025-nativqa,
    title = "{N}ativ{QA}: Multilingual Culturally-Aligned Natural Query for {LLM}s",
    author = "Hasan, Md. Arid  and
      Hasanain, Maram  and
      Ahmad, Fatema  and
      Laskar, Sahinur Rahman  and
      Upadhyay, Sunaya  and
      Sukhadia, Vrunda N  and
      Kutlu, Mucahid  and
      Chowdhury, Shammur Absar  and
      Alam, Firoj",
    editor = "Che, Wanxiang  and
      Nabende, Joyce  and
      Shutova, Ekaterina  and
      Pilehvar, Mohammad Taher",
    booktitle = "Findings of the Association for Computational Linguistics: ACL 2025",
    month = jul,
    year = "2025",
    address = "Vienna, Austria",
    publisher = "Association for Computational Linguistics",
    url = "https://aclanthology.org/2025.findings-acl.770/",
    doi = "10.18653/v1/2025.findings-acl.770",
    pages = "14886--14909",
    ISBN = "979-8-89176-256-5"
}

@inproceedings{myung2024blend,
  title={{BLEnD}: A Benchmark for LLMs on Everyday Knowledge in Diverse Cultures and Languages},
  author={Myung, Junho and Lee, Nayeon and Zhou, Yi and Jin, Jiho and Putri, Rifki Afina and Antypas, Dimosthenis and Borkakoty, Hsuvas and Kim, Eunsu and Perez-Almendros, Carla and Ayele, Abinew Ali and others},
  booktitle={Proceedings of the 38th Conference on Neural Information Processing Systems (NeurIPS)},
  year={2024},
  address={Vancouver, Canada},
  month={December}
}

@article{pawar2024survey,
  title={Survey of cultural awareness in language models: Text and beyond},
  author={Pawar, Siddhesh and Park, Junyeong and Jin, Jiho and Arora, Arnav and Myung, Junho and Yadav, Srishti and Haznitrama, Faiz Ghifari and Song, Inhwa and Oh, Alice and Augenstein, Isabelle},
  journal={Computational Linguistics},
  pages={1--96},
  year={2025},
  publisher={MIT Press}
}

@inproceedings{NEURIPS2024_77f089cd,
 author = {Li, Cheng and Teney, Damien and Yang, Linyi and Wen, Qingsong and Xie, Xing and Wang, Jindong},
 booktitle = {Advances in Neural Information Processing Systems},
 editor = {A. Globerson and L. Mackey and D. Belgrave and A. Fan and U. Paquet and J. Tomczak and C. Zhang},
 pages = {65183--65216},
 title = {{CulturePark}: Boosting Cross-cultural Understanding in Large Language Models},
 volume = {37},
 year = {2024}
}

@techreport{bubeck2023sparks,
      title={Sparks of Artificial General Intelligence: Early experiments with {GPT}-4}, 
      author={Sébastien Bubeck and Varun Chandrasekaran and Ronen Eldan and Johannes Gehrke and Eric Horvitz and Ece Kamar and Peter Lee and Yin Tat Lee and Yuanzhi Li and Scott Lundberg and Harsha Nori and Hamid Palangi and Marco Tulio Ribeiro and Yi Zhang},
      year={2023},
      eprint={2303.12712},
      archivePrefix={arXiv},
      primaryClass={cs.CL},
      institution={Microsoft Research}
}

@inproceedings{weichain,
  title={Chain-of-Thought Prompting Elicits Reasoning in Large Language Models},
  author={Wei, Jason and Wang, Xuezhi and Schuurmans, Dale and Bosma, Maarten and Xia, Fei and Chi, Ed H and Le, Quoc V and Zhou, Denny and others},
  booktitle={Advances in Neural Information Processing Systems},
  year={2022}
}

@inproceedings{antoun-etal-2020-arabert,
    title = "{A}ra{BERT}: Transformer-based Model for {A}rabic Language Understanding",
    author = "Antoun, Wissam  and
      Baly, Fady  and
      Hajj, Hazem",
    editor = "Al-Khalifa, Hend  and
      Magdy, Walid  and
      Darwish, Kareem  and
      Elsayed, Tamer  and
      Mubarak, Hamdy",
    booktitle = "Proceedings of the 4th Workshop on Open-Source Arabic Corpora and Processing Tools, with a Shared Task on Offensive Language Detection",
    month = may,
    year = "2020",
    address = "Marseille, France",
    publisher = "European Language Resource Association",
    url = "https://aclanthology.org/2020.osact-1.2",
    pages = "9--15",
    language = "English",
    ISBN = "979-10-95546-51-1",
}

% \section{Language Resource References}
% \label{lr:ref}
% \bibliographystylelanguageresource{lrec2026-natbib}
% \bibliographylanguageresource{languageresource}

\section{Appendix}

\subsection{Annotation Guidelines}
\label{sec:annotation_guidelines}

To ensure the quality and reliability of the generated dataset, we conducted structured human annotation and post-editing using an in-house annotation platform. This appendix provides an overview of the annotation setup and summarizes the instructions provided to annotators.

\subsubsection{MCQ$\rightarrow$OEQ Annotation}

The entire MSA OEQ test set, generated by transforming MSA MCQs into open-ended questions, was reviewed by three independent annotators. Annotators were presented with both the original MCQ (including the question, choices, and correct answer) and the automatically generated OEQ pair consisting of a question and answer.

An example of the annotation interface used for this task is shown in Figure~\ref{fig:oeq_annotation}. Annotators evaluated the quality of both the transformed question and the generated answer.

Annotators were asked to assign scores on a five-point Likert scale for two aspects:

\begin{itemize}
\item \textbf{Open-Ended Question Quality (1--5):} evaluating clarity, naturalness in MSA, completeness, and whether the question is self-contained with appropriate scope.
\item \textbf{Open-Ended Answer Quality (1--5):} evaluating correctness relative to the MCQ answer, completeness, conciseness, and fluency.
\end{itemize}

Scores were interpreted as follows:

\begin{itemize}
\item 1: Poor
\item 2: Weak
\item 3: Acceptable
\item 4: Good
\item 5: Excellent
\end{itemize}

When a score below 4 was assigned, annotators were required to select at least one revision reason explaining the issue. These included categories such as:

\begin{itemize}
\item unclear or ambiguous question
\item grammatical errors
\item not self-contained
\item overly broad or overly narrow scope
\item incorrect or incomplete answer
\item speculative or unsupported content
\end{itemize}

Annotators could also provide optional comments for particularly difficult or ambiguous cases.

% The full detailed annotation guideline document used in this task is available at:
% \footnote{\url{https://docs.google.com/document/d/1EP5hslKGvhSFTgM3LP7yyV9BYGo1fraEQKYvaHAdCQc/edit}}

\begin{figure}[t]
\centering
\includegraphics[width=\linewidth]{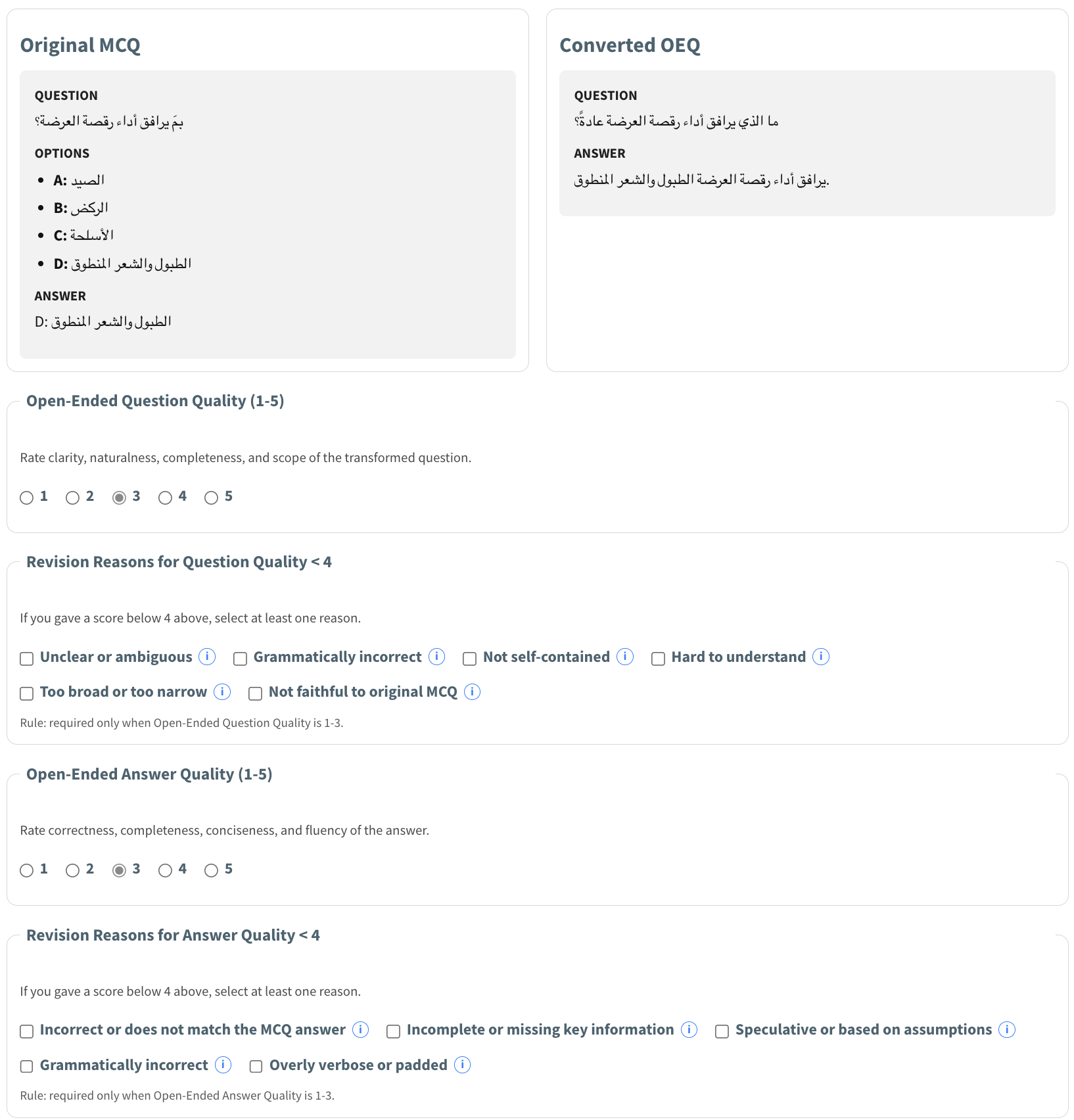}
\caption{Annotation interface for evaluating MCQ$\rightarrow$OEQ transformations. Annotators review the original MCQ alongside the generated open-ended question and answer.}
\label{fig:oeq_annotation}
\end{figure}

\subsubsection{Dialectal Translation Post-Editing}

For the dialectal datasets, we performed manual post-editing of the MCQ test sets after automatic translation. Each dialect subset was reviewed by a native speaker of the corresponding dialect who edited the translated question and answer options when necessary to improve naturalness and linguistic authenticity.

The annotation interface for this task is illustrated in Figure~\ref{fig:translation_annotation}. Annotators were shown:

\begin{itemize}
\item the original MSA content (read-only)
\item the translated version (editable)
\end{itemize}

Their task was to revise only the translated fields to ensure:

\begin{itemize}
\item semantic fidelity to the MSA source
\item fluent and natural wording in the target variety
\item consistent terminology and phrasing across question and options
\end{itemize}

Annotators were instructed not to modify the MSA fields and to keep the meaning faithful to the source question. Optional comments could be provided for unclear or difficult cases.

% The detailed translation editing guideline document is available at:
% \footnote{\url{https://docs.google.com/document/d/1BGuACx-gWHF5a6sCg_5SHOJ-2R5G2FvF/edit}}

\begin{figure}[t]
\centering
\includegraphics[width=\linewidth]{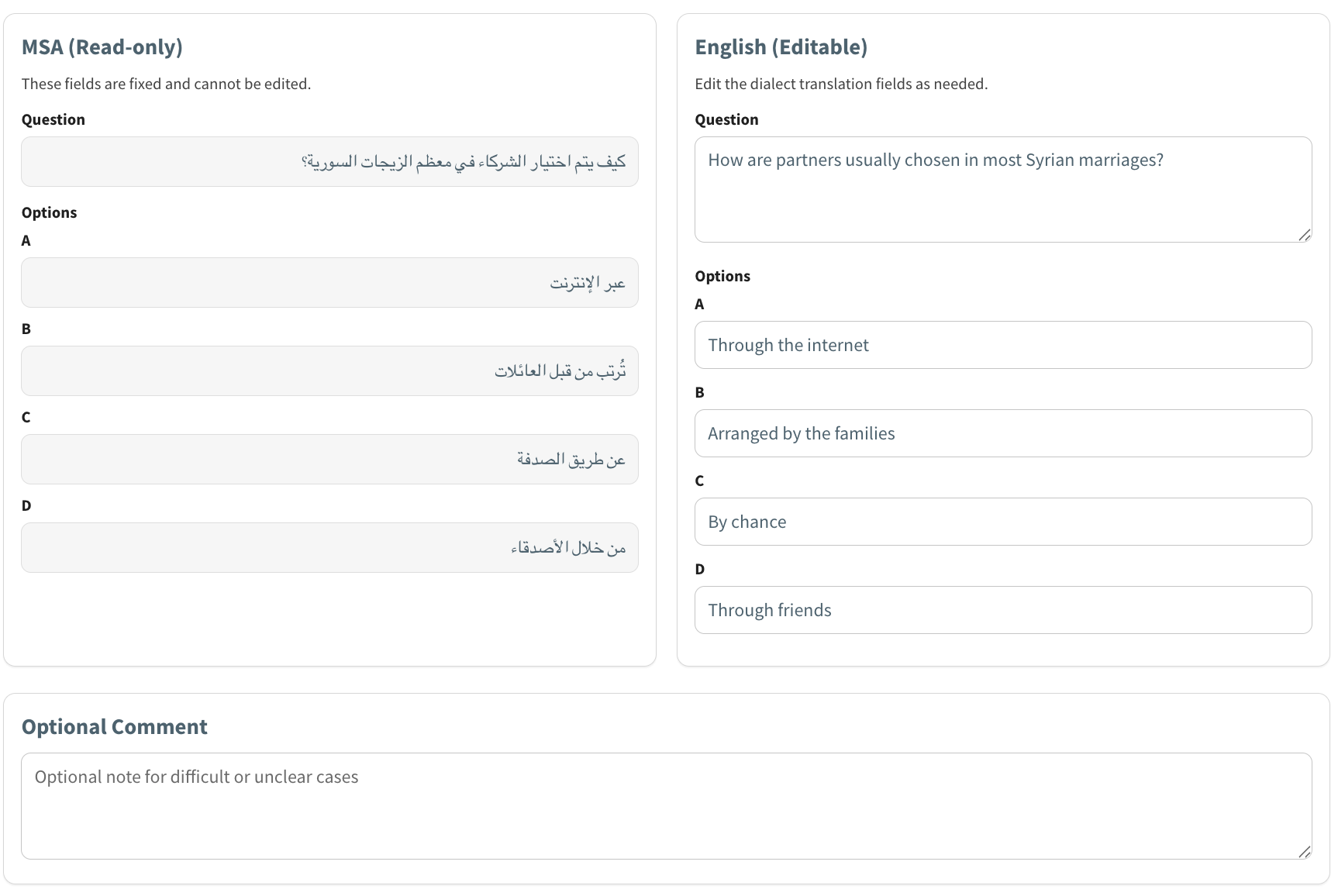}
\caption{Interface used for dialectal translation post-editing. Annotators review the original MSA question and revise the translated English version.}
\label{fig:translation_annotation}
\end{figure}

Together, these annotation and post-editing steps ensure that both the open-ended question transformations and the dialectal translations maintain semantic fidelity, linguistic quality, and cultural authenticity.

\subsection{Prompts}

This section provides the prompts used throughout this work, including those for chain-of-thought (CoT) reasoning generation, dataset construction, and benchmarking.

\subsubsection{Prompts for CoT Generation}
Listings 1--3 include the Solving, Rationalizing, and Verification prompts used to generate CoT reasoning.

\begin{figure*}[t]
\begin{lstlisting}[language={}, caption={Solve Prompt}]
[System Prompt]
You are a culturally knowledgeable assistant. Your task is to answer questions based on cultural context, history, and common perspectives.
1. First, analyze the question to determine if it's factual or subjective (opinion-based).
2. Provide step-by-step reasoning that explains the answer.
   - For factual questions, use established cultural or historical facts.
   - For subjective questions, explain the common viewpoints or the reasons behind a popular opinion, acknowledging its subjective nature.
3. Keep the reasoning concise and relevant.
4. Finally, provide a brief, direct answer.

Return ONLY a strict JSON object:
{"cot": "<step-by-step reasoning>", "final": "<short answer>"}

[User Prompt]
Question:
{question}

Respond as JSON only:
{"cot": "...", "final": "..."}.
\end{lstlisting}
% \end{figure*}

% \begin{figure*}[t]
\begin{lstlisting}[language={}, caption={Rationalization Prompt}]
[System Prompt]
You are a cultural explainer. Given a question and a known answer, your task is to construct a step-by-step explanation that coherently leads to that answer.
- The explanation must be grounded in relevant cultural knowledge, history, or common perceptions.
- For subjective answers, the reasoning should justify why this is a plausible or common perspective.
- Do not simply state the answer in the first step; build up to it through your explanation.

Return ONLY a strict JSON object:
{"cot": "<step-by-step reasoning>", "final": "<the provided answer>"}

[User Prompt]
Question:
{question}

Known correct answer:
{gold}

Construct an explanation that leads to the known answer. Restate the given answer in the 'final' field.
Respond as JSON only.
\end{lstlisting}
\end{figure*}

\begin{figure*}[t]
\begin{lstlisting}[language={}, caption={Verification Prompt}]
[System Prompt]
You are a meticulous cultural context verifier. Your job is to assess a proposed chain-of-thought (CoT) against a question and a gold answer.

Evaluate the CoT based on the following criteria:
1. Cultural Soundness: Is the reasoning culturally relevant and sound?
2. Factual Accuracy: Are factual claims correct?
3. Subjectivity Handling: If the question is subjective, does the CoT frame it as opinion rather than objective fact?
4. Logical Support: Does the reasoning support the final answer?

Provide a score from 0.0 to 1.0. The verdict is "pass" if score >= 0.8, otherwise "fail".
Return ONLY JSON:
{"score": <float>, "verdict": "pass|fail", "issues": ["..."]}

[User Prompt]
Question: {question}
Gold answer: {gold}
Proposed chain-of-thought (cot): {cot}
Proposed final answer: {final}

Assess the proposed CoT and final answer.
\end{lstlisting}
\end{figure*}

\subsubsection{Prompts for Dataset Construction and Evaluation}
\label{sec:other_prompts}
Listings 4--5 cover dataset construction (MCQ-to-MSA and dialectal conversion). Listings 6--7 provide the zero-shot MCQ and OEQ templates. Listings 8--9 contain the LLM-as-a-Judge prompt and output schema.

\begin{figure*}[t]
\begin{lstlisting}[language={}, caption={MCQ to MSA Open-Ended Conversion Prompt}]
You are given multiple-choice questions in JSON format. Your task is to convert them into open-ended Q&A format when possible.

Language: Always write the output in Modern Standard Arabic (MSA).

Rules:
- Rephrase the question so it works without multiple-choice options.
- Rephrase the correct answer as a natural full sentence.
- Use only the correct option for the final answer.
- If the question cannot reasonably be converted, mark it as not_possible.

Output format:
{
 "id": <same as input>,
 "open_question": "<MSA question>",
 "open_answer": "<MSA answer>",
 "status": "ok" | "not_possible"
}
\end{lstlisting}
\end{figure*}

\begin{figure*}[t]
\begin{lstlisting}[language={}, caption={MSA Open-Ended to Dialectal Conversion Prompt}]
You are a professional linguistic converter specializing in Arabic dialect adaptation.

Convert a Modern Standard Arabic question-answer pair into:
1. Gulf Arabic
2. Egyptian Arabic
3. Levantine Arabic
4. Maghrebi Arabic
5. English

Guidelines:
- Preserve meaning and cultural context
- Use natural dialect phrasing
- Maintain respectful tone
- Do not translate word-for-word

Return strictly JSON:

{
 "id": <same as input>,
 "msa_question": "...",
 "msa_answer": "...",
 "gulf_question": "...",
 "gulf_answer": "...",
 "egyptian_question": "...",
 "egyptian_answer": "...",
 "levantine_question": "...",
 "levantine_answer": "...",
 "maghrebi_question": "...",
 "maghrebi_answer": "...",
 "english_question": "...",
 "english_answer": "..."
}
\end{lstlisting}
% \end{figure*}

% \begin{figure*}[t]
\begin{lstlisting}[language={}, caption={MCQ Zero-Shot Prompt}]
You are an AI assistant specialized in answering multiple-choice questions about Arab culture and society.

Instructions:
- Read the question carefully
- Consider all options
- Respond ONLY with the correct letter (A, B, C, or D)
- Do not include any explanation

Your response must be exactly one letter.
\end{lstlisting}
\end{figure*}

\begin{figure*}[t]
\vspace{-1em} % Squeezes space at the very top of the figure

% --- FIRST PROMPT ---
\begin{lstlisting}[language={}, caption={Open-Ended Zero-Shot Prompt}, basicstyle=\small\ttfamily]
You are an expert Arabic culture scholar.

Answer open-ended questions directly and concisely.
Responses should reflect accurate cultural, social, or geographical knowledge.

Return only the answer with no explanation or extra text.
\end{lstlisting}

\vspace{-1.5em} % NEGATIVE VSPACE: Pulls the second prompt up significantly

% --- SECOND PROMPT: Split into 2 columns using minipages ---
\begin{minipage}[t]{0.48\textwidth}
\begin{lstlisting}[language={}, caption={LLM-as-Judge Prompt for OEQ Evaluation -- Instructions}, basicstyle=\small\ttfamily]
You are a strict, reference-grounded QA evaluator. Grade a single predicted
answer to an open-ended question only using the provided materials
(question, reference answer(s), and optional evidence/context).
Do NOT use outside knowledge. Judge in the same language as the
predicted answer (Arabic or English).

Return ONLY a valid JSON object that matches the schema exactly-no
extra text.

Score each criterion as an integer from 1 to 10 (1 = poor, 10 = excellent):

- Clarity: Is the writing easy to read, well-structured, and unambiguous?
- Informativeness: Does it cover the important points the question asks for?
- Plausibility: Is it internally coherent and reasonable given the question
  and references/context?
- Faithfulness: Are claims supported by the reference/evidence?
  No hallucinations or contradictions.
- Helpfulness: Does it directly address the user's need without fluff or evasion?
- Accuracy: Are facts correct relative to the reference/evidence
  (numbers, entities, relations)?
- Depth and Creativity: Nuance or insightful distinctions while remaining correct.
- Level of Detail: Appropriate specificity for the question.

Also compute an overall score (integer 1-10), typically the rounded mean
of the eight criteria.

Detect and list any unsupported claims, missing key points, or contradictions.
Be concise in rationales.

{Output_Schema}
\end{lstlisting}
\end{minipage}\hfill
\begin{minipage}[t]{0.48\textwidth}
\begin{lstlisting}[language={}, caption={LLM-as-Judge Prompt -- Output Schema}, basicstyle=\small\ttfamily]
Output_Schema:
{
  "scores": {
    "clarity": 1,
    "informativeness": 1,
    "plausibility": 1,
    "faithfulness": 1,
    "helpfulness": 1,
    "accuracy": 1,
    "depth_creativity": 1,
    "level_of_detail": 1,
    "overall": 1
  },
  "rationale": {
    "overall": "1-3 sentences justifying the overall score.",
    "faithfulness": "Key supported/unsupported points in 1-3 sentences.",
    "accuracy": "Any factual errors vs. reference/evidence in 1-3 sentences."
  },
  "findings": {
    "unsupported_claims": ["..."],
    "missing_key_points": ["..."],
    "contradictions": ["..."]
  }
}
\end{lstlisting}
\end{minipage}

\vspace{-1em} % Squeezes space at the bottom before the main document text resumes
\end{figure*}

\begin{figure*}
    \centering
    \includegraphics[width=\linewidth]{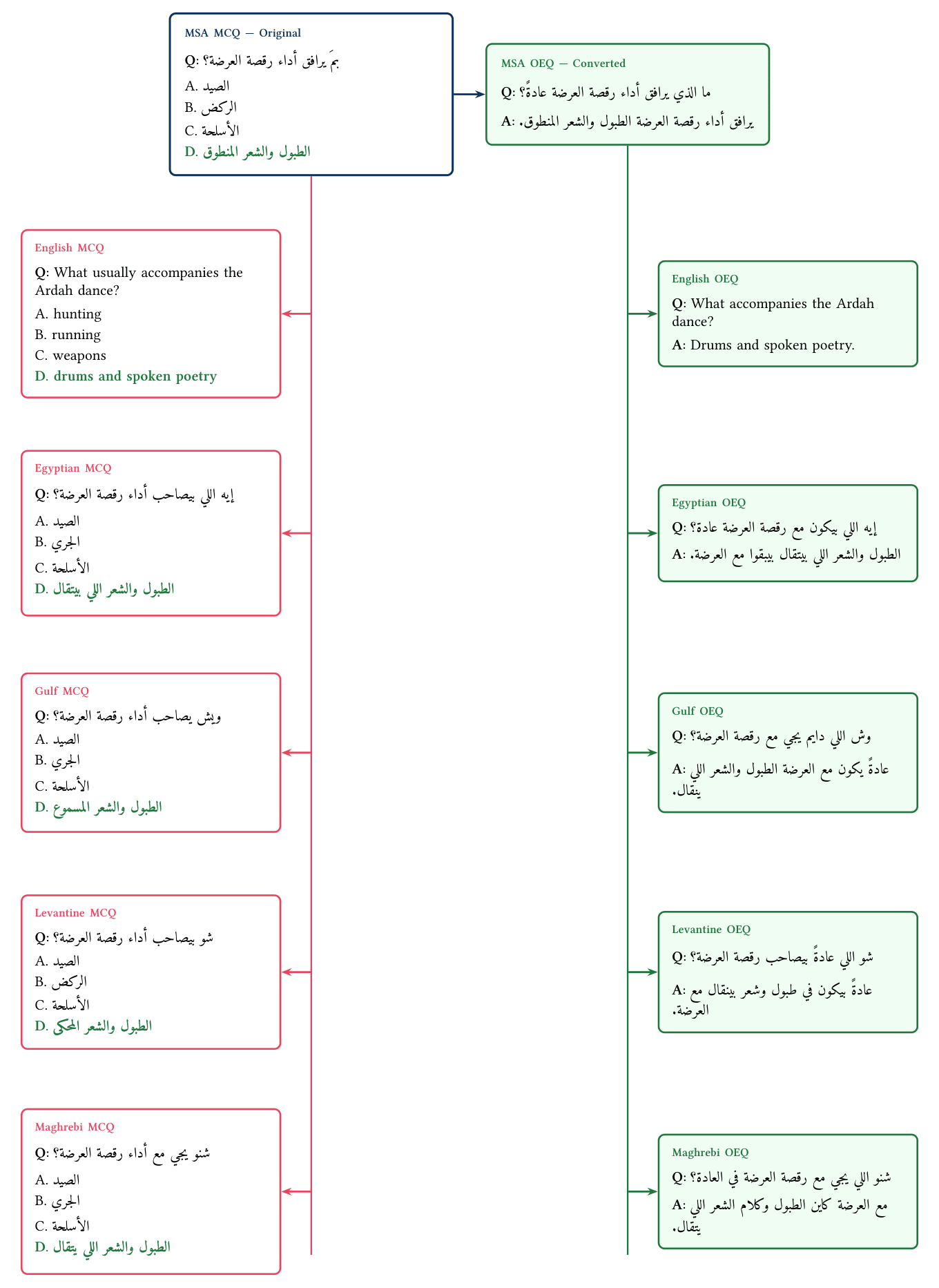}
    \caption{An example from the generated test set before post-editing and annotation. The MSA MCQ is first converted into the target dialects and simultaneously transformed into an MSA OEQ. The resulting MSA OEQ is then further adapted into its corresponding dialectal OEQ variants.}
    \label{fig:example_diagram}
\end{figure*}

\end{document}